\documentclass[pmlr]{jmlr}

\usepackage{longtable}% for long tables

\usepackage{booktabs}
\usepackage[load-configurations=version-1]{siunitx} % newer version

\usepackage{amssymb}
\usepackage{amsmath}
\usepackage{pifont}
\usepackage{float}
\usepackage{seqsplit}
\usepackage{cancel}

\newcommand{\cmark}{\ding{51}}%
\newcommand{\xmark}{\ding{55}}%

\theorembodyfont{\upshape}
\theoremheaderfont{\scshape}
\theorempostheader{:}
\theoremsep{\newline}

\jmlrvolume{}
\firstpageno{1}
\editors{Sophia Sanborn, Christian Shewmake, Simone Azeglio, Nina Miolane}

\jmlryear{2023}
\jmlrworkshop{Symmetry and Geometry in Neural Representations}

\title[Homological Convolutional Neural Networks]{Homological Convolutional Neural Networks}

\author{\Name{Antonio Briola} \Email{antonio.briola.20@ucl.ac.uk}\\
\Name{Yuanrong Wang} \Email{yuanrong.wang.20@ucl.ac.uk}\\
\Name{Silvia Bartolucci} \Email{s.bartolucci@ucl.ac.uk}\\
\Name{Tomaso Aste} \Email{t.aste@ucl.ac.uk}\\
\addr Department of Computer Science, University College London, London, WC1E 6BT, UK}

\begin{document}
\pagenumbering{gobble}

\maketitle

\begin{abstract}
Deep learning methods have demonstrated outstanding performances on classification and regression tasks on homogeneous data types (e.g., image, audio, and text data). However, tabular data still pose a challenge, with classic machine learning approaches being often computationally cheaper and equally effective than increasingly complex deep learning architectures. The challenge arises from the fact that, in tabular data, the correlation among features is weaker than the one from spatial or semantic relationships in images or natural language, and the dependency structures need to be modeled without any prior information. In this work, we propose a novel deep learning architecture that exploits the data structural organization through topologically constrained network representations to gain relational information from sparse tabular inputs. The resulting model leverages the power of convolution and is centered on a limited number of concepts from network topology to guarantee: (i) a data-centric and deterministic building pipeline; (ii) a high level of interpretability over the inference process; and (iii) an adequate room for scalability. We test our model on $18$ benchmark datasets against $5$ classic machine learning and $3$ deep learning models, demonstrating that our approach reaches state-of-the-art performances on these challenging datasets. The code to reproduce all our experiments is provided at \url{https://github.com/FinancialComputingUCL/HomologicalCNN}.
\end{abstract}

\section{Introduction} \label{sec:Introduction}

We are experiencing a tremendous and inexorable progress in the field of deep learning. Such a progress has been catalyzed by the availability of increasing computational resources and always larger datasets. The areas of success of deep learning are heterogeneous. However, the three application domains where superior performances have been detected are the ones involving the usage of images \citep{he2015delving, pak2017review}, audio \citep{purwins2019deep, bose2020deep} and text \citep{lai2015recurrent, chowdhary2020natural, zhang2021commentary}. Despite their inherent diversity, these data types share a fundamental characteristic: they exhibit homogeneity, with notable inter-feature correlations and evident spatial or semantic relationships. On the contrary, tabular data represent the “unconquered castle” of deep neural network models \citep{kadra2021well}. They are heterogeneous data types and present a mixture of continuous, categorical, and ordinal values, which can be either independent or correlated. They are characterized by the absence of any inherent positional information, and tabular models have to handle features from multiple discrete and continuous distributions. However, tabular data are the most common data format and are ubiquitous in many critical applications, such as medicine \citep{ulmer2020trust, somani2021deep}, finance \citep{sachan2020explainable, ohana2021explainable}, recommendation systems \citep{zhang2019deep, zhang2021neural}, cybersecurity \citep{buczak2015survey, rawat2019cybersecurity}, and anomaly detection \citep{pang2022editorial, wang2022multiview} -- to mention a few. During the last decade, traditional machine learning methods dominated tabular data modeling, and, nowadays, tree ensemble algorithms (i.e. XGBoost, LightGBM, CatBoost) are the recommended option to solve real-life problems of this kind \citep{friedman2001greedy, prokhorenkova2018catboost, shwartz2022tabular}. 

In the current paper, we introduce a novel deep learning architecture for tabular numerical data classification and we name it ``Homological Convolutional Neural Network” (\verb|HCNN|). We exploit a class of information filtering networks \citep{barfuss2016parsimonious, briola2022anatomy, briola2022dependency, briola2023topological, vidal2023ftx, wang2023topological}, namely the Triangulated Maximally Filtered Graph \citep{massara2017network}, to model the inner sparsity of tabular data and obtain a geometrical organization of input features. Emerging data relationships are hence studied at different granularity levels to capture both simplicial and homological structures through the usage of Convolutional Neural Networks (CNNs). Compared to state-of-the-art (SOTA) machine learning alternatives \citep{friedman2001greedy, chen2016xgboost, ke2017lightgbm, prokhorenkova2018catboost}, our method (i) maintains an equivalent level of explainability; (ii) has a comparatively lower level of computational complexity; and (iii) can be scaled to a higher number of learning tasks (e.g. time series forecasting) without structural changes. Compared to its SOTA deep-learning alternatives \citep{arik2021tabnet, badirli2020gradient, hazimeh2020tree, huang2020tabtransformer, klambauer2017self, kontschieder2015deep, popov2019neural, song2019autoint, somepalli2021saint, beutel2018latent, wang2019learning, kadra2021regularization, kadra2021well, shavitt2018regularization, baosenguo2021}, our method (i) is data-centric (i.e. the architecture depends on the data defining the system under analysis); (ii) presents an algorithmic data-driven building pipeline; and (iii) has a lower complexity, replacing complex architectural modules (e.g. attention-based mechanisms) with elementary computational units (e.g. convolutional layers). We provide a comparison between \verb|HCNN|s, simple-to-advanced machine learning algorithms and SOTA deep tabular architectures using a heterogeneous battery of small-to-medium sized numerical benchmark datasets. We observe that \verb|HCNN| always ties SOTA performances on the proposed tasks, providing, at the same time, structural and computational advantages.

\section{Data and Methods} 
\subsection{Data} \label{sec:Data}
To provide a fair comparison between \verb|HCNN| and SOTA models, we use a collection of $18$ tabular numerical datasets from the open-source “OpenML-CC18” benchmark suite \citep{bischl2017openml}. Following the selection criteria in \citep{hollmann2022tabpfn}, all the datasets contain up to $2\,000$ samples, $100$ features, and $10$ classes. A deep overview on the properties of this first set of data is provided in \appendixref{app:Appendix_Small_Tabular_Data}. Following \citep{grinsztajn2022tree}, we focus on small datasets because of two main reasons: (i) small datasets are often encountered in real-world applications \citep{dua2017uci}; and (ii) existing deep learning methods are limited in this domain. It is worth noticing that, differently from other deep learning architectures \citep{hollmann2022tabpfn, arik2021tabnet}, the applicability of \verb|HCNN|s is not limited to small tabular data problems and can easily scale to medium-to-large problems. To provide evidence of this, we use a collection of $9$ numerical tabular datasets from the “OpenML tabular benchmark numerical classification” suite \citep{grinsztajn2022tree}. All these datasets violate at least one of the selection criteria in \citep{hollmann2022tabpfn} (i.e. they are characterized by a number of samples $> 2\,000$ or they are characterized by a number of features $> 100$). A deep overview on the properties of this second set of data is provided in \appendixref{app:Appendix_Small_Tabular_Data}.

\subsection{Information Filtering Networks}\label{sec:Information_Filtering_Networks}
The \verb|HCNN|'s building process is entirely centered on the structural organization of data emerging from the underlying sparse network representation. The choice of the network representation is not binding even if limited to the family of simplicial complexes \citep{torres2020simplicial, salnikov2018simplicial}. In this paper, we exploit the power of a class of information filtering networks (IFNs) \citep{mantegna1999hierarchical, aste2005complex, barfuss2016parsimonious, massara2017network, tumminello2005tool}, namely the Triangulated Maximally Filtered Graph (TMFG) \citep{massara2017network}, to model the inner sparsity of tabular data and obtain a structural organization of input features. IFNs are an effective tool to represent and model dependency structures among variables characterizing complex systems while imposing topological constraints (e.g. being a tree or a planar graph) and optimizing specific global properties (e.g. the likelihood) \citep{aste2022topological}. Starting from a system characterized by $n$ features and $T$ samples, arranged in a matrix \textbf{X}, this methodology builds a $n \times n$ similarity matrix $\hat{\textbf{C}}$ which is filtered to obtain a sparse adjacency matrix \textbf{\textit{A}} retaining only the most structurally significant relationships among variables. Working with numerical-only tabular data, in the current paper, $\hat{\textbf{C}}$ corresponds to a matrix of squared correlation coefficients. To improve the robustness of the correlation similarity measure, in line with the work by \citep{tumminello2007spanning}, we use the bootstrapping approach \citep{efron1996bootstrap}. This technique requires to build a number $r$ of replicas $X_{i}^{*}$, $i \in 1, \dots, r$ of the data matrix \textbf{X}. Each replica $X_{i}^{*}$ is built by randomly selecting $T$ rows from the matrix \textbf{X} allowing for repetitions. For each replica $X_{i}^{*}$, the correlation matrix $\hat{\textbf{C}}_{i}^{*}$ is then computed. We highlight that (i) the bootstrap approach does not require the knowledge of the data distribution; and (ii) it is particularly useful to deal with high dimensional systems where it is difficult to infer the joint probability distribution from data. Once obtained replicas-dependent correlation matrices, we treat them in two different ways:

\begin{itemize}
    \item We compute $\hat{\textbf{C}}$ as the entry-wise mean of correlation matrices $\hat{\textbf{C}}_{i \in 1, \dots, r}^{*}$, and we construct a TMFG (see \appendixref{app:Appendix_TMFG_Algo}) by using it. 
    \item Based on each replica-dependent correlation matrix $\hat{\textbf{C}}_{i}^{*}$, we compute a $\textnormal{TMFG}_{i}^{*}$ (see \appendixref{app:Appendix_TMFG_Algo}) and we obtain the final, filtered, TMFG by taking only the links that appear in all the $\textnormal{TMFG}^{*}$ with a frequency higher than a specified threshold.
\end{itemize}

In the rest of the paper, we will refer to the first configuration as \verb|MeanSimMatrix| and to the second one as \verb|BootstrapNet|. These two approaches lead to widely different results. In the former case, the final TMFG will be a sparse, connected graph that necessarily maintains all the topological characterization of the family of IFNs it belongs to (i.e. planarity \citep{tumminello2005tool} and chordality \citep{massara2017network}). In the latter case, instead, there will be no guarantee on the connectedness of the graph. Indeed, the chosen threshold could lead to disconnected components and to the removal of edges assuring the graph's chordality.

\subsection{Homological Convolutional Neural Networks}\label{sec:Homological_Convolutional_Neural_Networks} The main idea behind IFNs is to explicitly model higher-order sub-structures, which are crucial for the representation of the underlying system's interactions. Based on this, in a recent work \citep{wang2023homological}, the authors propose a simple higher-order representation which (i) starts from a layered representation (i.e. the Hasse diagram); (ii) explicitly takes into account higher-order sub-structures (i.e. simplices) and their higher-order interconnections (i.e. homological priors); (iii) and converts this representation into a stand-alone computational unit named ``Homological Neural Network” (\verb|HNN|). Despite the undeniable advantages deriving from this sparse architecture (see \appendixref{app:Appendix_HNN_HCNN}), results suggest that the authors' choice of using the Multilayer Perceptron as a deep learning architecture to process the information encoded in the underlying network representation, is sub-optimal (especially for tabular data problems). In addition to this, \verb|HNN|s impose the chordality of the underlying network and the building process of the deep neural network architecture implies the usage of non-native components inducing a substantial computational overhead while limiting its applicability to simple classification and regression problems. In this research work, we propose an alternative computational architecture that aims to solve these issues and we name it ``Homological Convolutional Neural Network” (\verb|HCNN|).

\begin{figure}[h]\label{HCNN_Solo_Schema}
    \vspace{-10pt}
    \centering
    \includegraphics[scale=0.2]{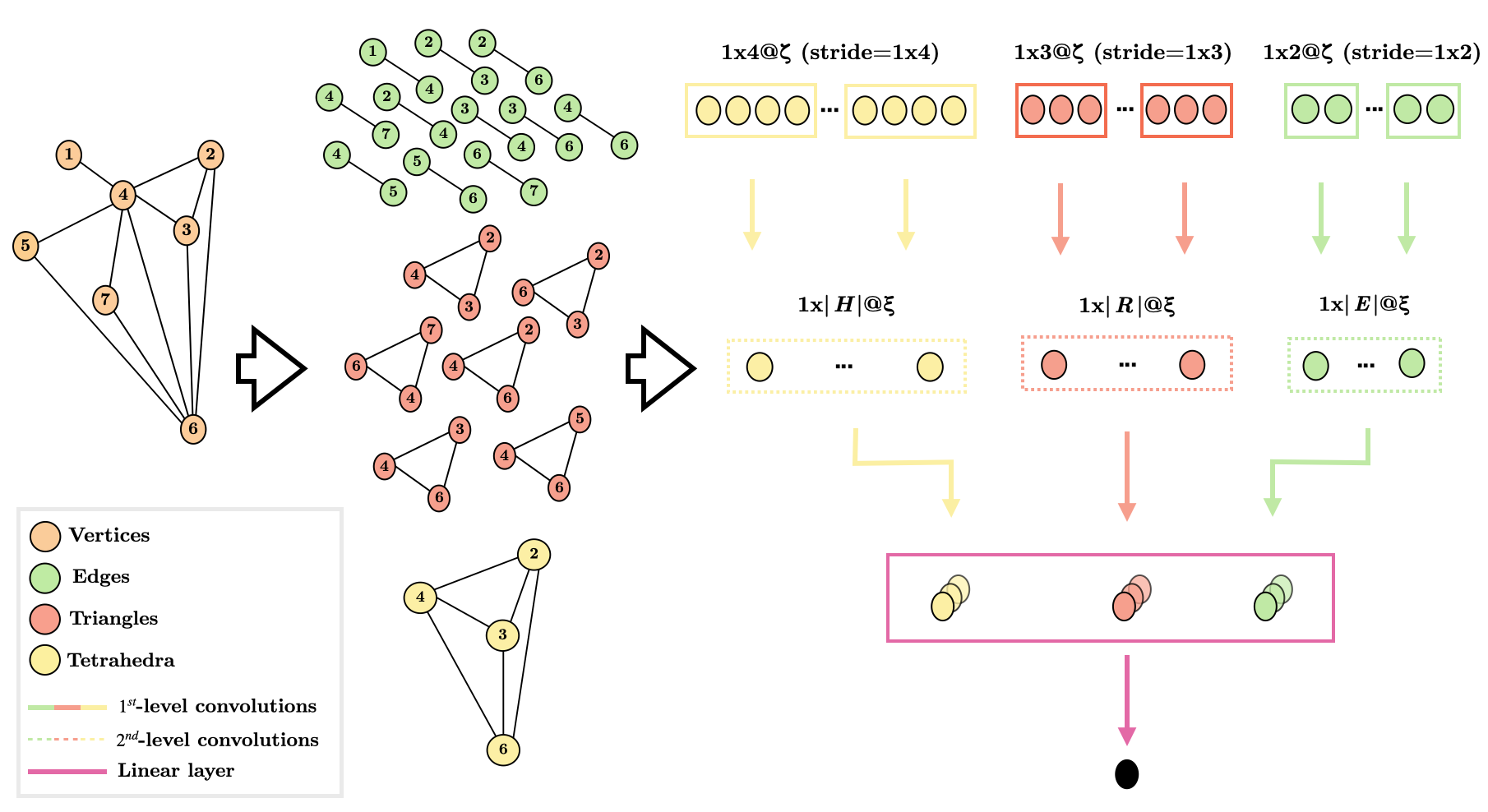}
    \caption{Pictorial representation of an HCNN and its building pipeline. From left to right, (i) we start from a chordal graph representing the dependency structures of features in the underlying system; (ii) we isolate the maximal cliques corresponding to $1$-, $2$- and $3$-dimensional simplices (i.e. edges, triangles, tetrahedra) and we group them into $1$D vectors containing features' realizations; (iii) we compute a $1^{st}$-level convolution to extract simplicial-wise non-linear relationships; (iv) we compute a $2^{nd}$-level convolution, which operates on the output of the previous level of convolution across all the representatives of each simplicial family extracting a class of non-trivial homological insights; (v) we finally apply a linear map from the $2^{nd}$-level convolution to the output, extracting model's predictions.}
    \label{fig:Solo_HCNN}
\end{figure}

Given the adjacency matrix \textbf{\textit{A}} constructed using IFNs (see \sectionref{sec:Information_Filtering_Networks}), 
we isolate $3$ different simplicial families: (i) maximal cliques with size $4$ (i.e. $3$-dimensional simplices or tetrahedra); (ii) maximal cliques with size $3$ (i.e. $2$-dimensional simplices or triangles); and (iii) maximal cliques with size $2$ (i.e. $1$-dimensional simplices or edges). When using the TMFG as a network representation, these $3$ structures are sufficient to capture all the higher-order dependency structures characterizing the underlying system. The input of the novel deep learning architecture is hence represented by $3$ different $1$D vectors that we call $H$ (i.e. realizations of the input features belonging to at least one tetrahedron), $R$ (i.e. realizations of the input features belonging to at least one triangle), and $E$ (i.e. realizations of the input features belonging to at least one edge) respectively. As a first step, in \verb|HCNN|, we perform a $1$D convolution across each set of features defining a realization of a simplicial family. We use a kernel size and a stride equals to $k+1$ (i.e. the dimension of the simplicial structure itself), and a number of filters $\zeta \in [4, 8, 12, 16]$ (notice that, as described in \sectionref{sec:Experiments}, the punctual value is chosen through an extended hyper-parameters search). This means that, given the three input vectors $H$, $R$ and $E$ representing the three simplicial families characterizing a TMFG, we compute a $1$D convolution with a kernel size and a stride equal to $2$, $3$ and $4$ respectively for edges, triangles, and tetrahedra. The usage of stride is necessary to prevent the “parameter sharing” \citep{zhang2019deeplob}. While generally considered an attractive property as fewer parameters are estimated and overfitting is avoided, in our case it leads to inconsistencies. Indeed, geometrical structures belonging to the same simplicial family (i.e. edges, triangles, and tetrahedra), but independent in the hierarchical dependency structure of the system, would share parameters.

After the $1^{st}$-level convolution, which extract element-wise information from geometrical structures belonging to the same simplicial family, we apply a $2^{nd}$-level convolution to extract further homological insights. Indeed, the convolution is applied to the output of the first layer, extracting information related to entities belonging to the same simplicial family, which are not necessarily related in the original network representation. In this case, we use a kernel with a size equal to the cardinality of the simplicial family (i.e. $|E|$, $|R|$, $|H|$ respectively) and a number of filters $\xi \in [32-64]$ (notice that, also in this case, the punctual value is chosen through an extended hyper-parameters search). The final layer of the \verb|HCNN| architecture is linear, and maps the outputs from the $2^{nd}$-level convolution to the output. It is worth noticing that each level of convolution is followed by a regularization layer with a dropout rate equal to $0.25$ and the non-linear activation function is the standard Rectified Linear Unit (ReLU). As one can see, the \verb|HCNN|'s building pipeline is fully explainable: (i) we start from a network representation (i.e. the IFN) that captures the system's multivariate probability distribution by maximising its likelihood (see \sectionref{sec:Formal_justification}) and gaining relational information from sparse tabular inputs; and (ii) we transform such a representation into a standalone neural network architecture by mapping each topological prior into a computational block. The only hyper-parameter subject to tuning is the number of filters in both layers of the \verb|HCNN| (i.e. $\zeta$ and $\xi$); all the other architectural choices are deterministic and led by the topological structure of the IFN describing the underlying system.

\subsection{On the learning process of network's representation} \label{sec:Formal_justification}
According to the setup described in \sectionref{sec:Homological_Convolutional_Neural_Networks}, the architecture proposed in the current paper arises from a graph-based higher-order representation $\mathcal{G}$ of a multivariate system $\textbf{X} = (X_1, \dots, X_n)^\top$, where the components $X_i$ are unidimensional scalar random variables characterized by an (unknown) probability density function $f(\textbf{X})$. Our primary goal is hence estimating the multivariate probability density function with representation structure $\mathcal{G}^*$, $\widetilde{f}(\textbf{X}|\mathcal{G}^*)$, that best describes the true and unknown $f(\textbf{X})$. This problem is known to be NP-hard \citep{osman2003greedy}, however one can restrict the search space and identify a priori the optimal network representation by analysing the dependency structure of the features characterizing the system under analysis. From an information theoretic perspective, the learning of an optimal network representation $\mathcal{G}^*$ consists of minimising the Kullback-Leibler divergence ($D_{KL}$) \citep{kullback1951information} between $f(\textbf{X})$ and $\widetilde{f}(\textbf{X}|\mathcal{G})$, and, consequently, the cross-entropy ($H$) of the underlying system:
            
\begin{equation}\label{eq:new_general_problem_formulation}
    \begin{aligned}
        \mathcal{G}^* &\Rightarrow \arg\min_{\mathcal{G}}{D_{KL}(f(\textbf{X}) \;||\; \widetilde{f}(\textbf{X} |\mathcal{G}})) \\
    &\Rightarrow \arg\min_{\mathcal{G}}{\cancel{\mathbb{E}_f (\log f(\textbf{X}))} - \mathbb{E}_f (\log \widetilde{f}(\textbf{X} | \mathcal{G}))} \\
    &\Rightarrow \arg\min_{\mathcal{G}}(H(\textbf{X}|\mathcal{G}))
    \end{aligned}
\end{equation}

The term $\mathbb{E}_f (\log f(\textbf{X}))$ in \equationref{eq:new_general_problem_formulation} is independent of $\mathcal{G}$ and therefore its value is irrelevant to the purpose of discovering the optimal representation network. The second term, $- \mathbb E(\log \widetilde{f}(\mathbf X | \mathcal G))$ (notice the minus), instead depends on $ \mathcal G$ and must be minimized. It is the estimate of the entropy of the multivariate system under analysis and corresponds to the so-called cross-entropy. Given that the true underlying distribution is unknown, the expectation cannot be computed exactly, however, it can be estimated with arbitrary precision using the sample mean. Such a sample mean approximates the expected value of the negative log-likelihood of the model $\widetilde{f}(\mathbf{X} | \mathcal{G})$. Therefore, the previously described optimization problem, becomes a likelihood ($\mathcal{L}$) maximization problem. The network associated with the largest system's likelihood can be constructed step-by-step by joining disconnected parts that share the largest mutual information. Indeed, in a graph, the gain achieved by joining two variables $ X_a$ and $X_b$, is approximately given by the mutual information shared by the two variables $\simeq  I( X_a; X_b)$. In turn, at the second-order approximation, the mutual information corresponds to the squared correlation coefficient between the two variables. It follows that the gain in likelihood is $I( X_a; X_b) \simeq \rho_{a,b}^2$ \citep{ihara1993information}, and the TMFG construction with $\rho^2$ weights implies a graph that maximizes the system's likelihood itself.

\section{Experiments}\label{sec:Experiments}
In this section, we compare the performance of the \verb|HCNN| classifier in its \verb|MeanSimMatrix| and \verb|BootstrapNet| configuration (see \sectionref{sec:Information_Filtering_Networks}) against $5$ machine learning and $3$ deep learning SOTA classifiers under homogeneous evaluation conditions. We consider \verb|LogisticRegression|, \verb|RandomForest|, \verb|XGBoost|, \verb|LightGBM| and \verb|CatBoost| as representatives of machine learning classifiers, and \verb|MLP|, \verb|TabNet| and \verb|TabPFN| as representatives of deep learning classifiers. For each of them, the inference process is structured into two different phases: (i) the hyper-parameters search stage; and (ii) the training/test stage with optimal hyper-parameters. Both stages are repeated $10$ times with fixed seeds (see \appendixref{app:Appendix_Small_Tabular_Data}) to guarantee a full reproducibility of results. For each run, we allow for a maximum of $500$ hyper-parameters search iterations allocating $8$ CPUs with $2$GB memory each and a time budget of $48$ hours. Obtained results are statistically validated using the Wilcoxon significance test, a standard metric for comparing classifiers across multiple datasets \citep{demvsar2006statistical}. On a second stage of the analysis, we investigate the scalability of each model in tackling extensive numerical tabular classification tasks. In so doing, we use an ad-hoc suite of datasets (see \sectionref{sec:Data}), while maintaining the inference process described earlier in this Section. A model converges (i.e. it is able to scale to larger classification problems) once completing the learning task using the given computational resources in the allocated time budget for all the $10$ seeds.

\subsection{Small tabular classification problems} \label{sec:Small_tabular_classification_results}

\tableref{tab:General_Table_Small_Tabular_Classification_Problems} reports a cross-datasets, out-of-sample comparison of classifiers previously listed in this section. For each model, we provide (i) the average; (ii) the best/worst ranking position considering three different evaluation metrics; (iii) the average value for each evaluation metric;  (iv) the time required for the hyper-parameters tuning; and (v) for the training/test run with optimal hyper-parameters. An extended version of these results is provided in \appendixref{app:Appendix_Radar_Plots} and in \appendixref{app:Appendix_Dataset_Wise_Scores_Small_Tabular_Data}. 

\begin{table}[h]\label{tab:General_Table_Small_Tabular_Classification_Problems}
\vspace{-10pt}
\caption{Cross-datasets, out-of-sample comparison of classifiers' performance. For each model, we provide (i) the average (“M.” abbreviation); (ii) the best/worst (“B/W” abbreviation) ranking position considering three different evaluation metrics; (iii) the average (“M.” abbreviation) value for each evaluation metric; and (iv) the time in seconds ($s$) required for the hyper-parameters tuning and for the training/test run with optimal hyper-parameters.}
\centering
\resizebox{\columnwidth}{!}{%
\begin{tabular}{ccccccccccc}
\hline
 &
  \textbf{LogisticRegression} &
  \textbf{RandomForest} &
  \textbf{XGBoost} &
  \textbf{LightGBM} &
  \textbf{CatBoost} &
  \textbf{MLP} &
  \textbf{TabNet} &
  \textbf{TabPFN} &
  \textbf{\begin{tabular}[c]{@{}c@{}}HCNN\\ BootstrapNet\end{tabular}} &
  \textbf{\begin{tabular}[c]{@{}c@{}}HCNN\\ MeanSimMatrix\end{tabular}} \\ \hline
\textbf{M. rank F1\_Score} &
  5.333 &
  6.333 &
  5.500 &
  5.277 &
  4.666 &
  9.500 &
  7.388 &
  2.388 &
  4.888 &
  3.722 \\
\textbf{M. rank Accuracy} &
  4.888 &
  5.972 &
  6.194 &
  5.916 &
  5.388 &
  9.666 &
  7.166 &
  1.833 &
  4.694 &
  3.277 \\
\textbf{M. rank MCC} &
  5.166 &
  6.388 &
  5.611 &
  5.666 &
  4.833 &
  9.500 &
  7.333 &
  2.166 &
  4.777 &
  3.5556 \\ \hline
\textbf{B/W rank F1\_Score} &
  1-9 &
  1-10 &
  1-9 &
  1-8 &
  1-8 &
  8-10 &
  2-10 &
  1-10 &
  1-9 &
  2-7 \\
\textbf{B/W rank Accuracy} &
  1-10 &
  1-10 &
  3-9 &
  2-8 &
  1-10 &
  8-10 &
  2-10 &
  1-5 &
  1-9 &
  1-6 \\
\textbf{B/W rank MCC} &
  1-10 &
  2-10 &
  1-9 &
  1-8 &
  1-8 &
  8-10 &
  2-10 &
  1-8 &
  1-9 &
  2-6 \\ \hline
\textbf{M. F1\_Score} &
  0.79$\pm$0.14 &
  0.77$\pm$0.15 &
  0.81$\pm$0.13 &
  0.80$\pm$0.12 &
  0.80$\pm$0.12 &
  0.74$\pm$0.15 &
  0.77$\pm$0.15 &
  0.83$\pm$0.14 &
  0.80$\pm$0.14 &
  0.81$\pm$0.13 \\
\textbf{M. Accuracy} &
  0.86$\pm$0.09 &
  0.86$\pm$0.09 &
  0.87$\pm$0.10 &
  0.86$\pm$0.09 &
  0.86$\pm$0.10 &
  0.83$\pm$0.09 &
  0.86$\pm$0.09 &
  0.88$\pm$0.09 &
  0.87$\pm$0.09 &
  0.88$\pm$0.09 \\
\textbf{M. MCC} &
  0.68$\pm$0.24 &
  0.65$\pm$0.27 &
  0.69$\pm$0.24 &
  0.68$\pm$0.23 &
  0.68$\pm$0.24 &
  0.59$\pm$0.29 &
  0.64$\pm$0.28 &
  0.71$\pm$0.26 &
  0.68$\pm$0.26 &
  0.70$\pm$0.25 \\ \hline
\textbf{M. Time (s) (Tune)} &
  22.95 &
  3097.83 &
  842.08 &
  900.24 &
  3704.13 &
  136.47 &
  37148.33 &
  11475.86 &
  6349.97 &
  7103.76 \\
\textbf{M. Time (s) (Train + Test)} &
  0.03 &
  1.24 &
  0.88 &
  4.49 &
  17.73 &
  0.17 &
  35.82 &
  15.51 &
  17.09 &
  15.58 \\ \hline
\end{tabular}%
}
\label{tab:Small_Tabular_Classification_Compact_Results}
\end{table}

From the table above, one can observe that, on average, the \verb|TabPFN| model occupies a ranking position higher than the one of \verb|HCNN| both in its \verb|MeanSimMatrix| and in \verb|BootstrapNet| configuration. However, it is worth noticing that when we evaluate models' performance through the F1\_Score and the Matthew's Correlation Coefficient (MCC) (i.e. the two performance metrics that are less prone to bias induced by unbalanced datasets), the \verb|HCNN| in the \verb|MeanSimMatrix| configuration occupies a ranking position for the worst performance which is better than the one of its immediate competitor (i.e. position $7$ and $6$ for \verb|HCNN MeanSimMatrix| vs position $10$ and $8$ for \verb|TabPFN|). The same happens in the case of the \verb|HCNN BootstrapNet| with the F1\_score. These findings highlight an evident robustness of the \verb|HCNN| model, which is superior not only to \verb|TabPFN| model, but also to all the other deep learning and machine learning alternatives. More generally, both \verb|TabPFN| and \verb|HCNN| show superior performances compared to the other two deep learning models (i.e. \verb|MLP| and \verb|TabNet|), which occupy an average ranking position equal to $\sim7$ and $\sim9$ respectively for all the three different evaluation metrics. Among machine learning models, \verb|CatBoost| achieves the highest performance with an average ranking position equal to $\sim4$ considering the F1\_Score and the MCC, and equal to $\sim5$ considering the accuracy score (in this case the position number $4$ is occupied by \verb|LogisticRegression|).

\begin{figure}[h]
    \vspace{-10pt}
    \centering
    \subfigure[]{\includegraphics[scale=0.43]{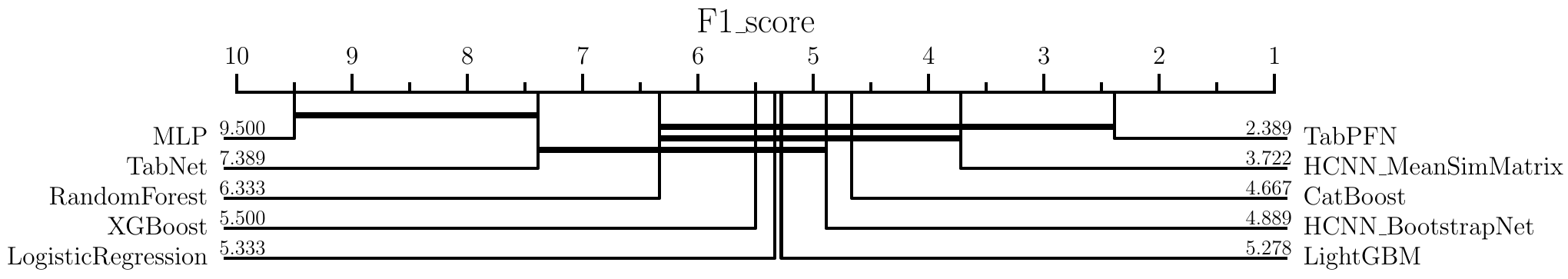}}
    \subfigure[]{\includegraphics[scale=0.43]{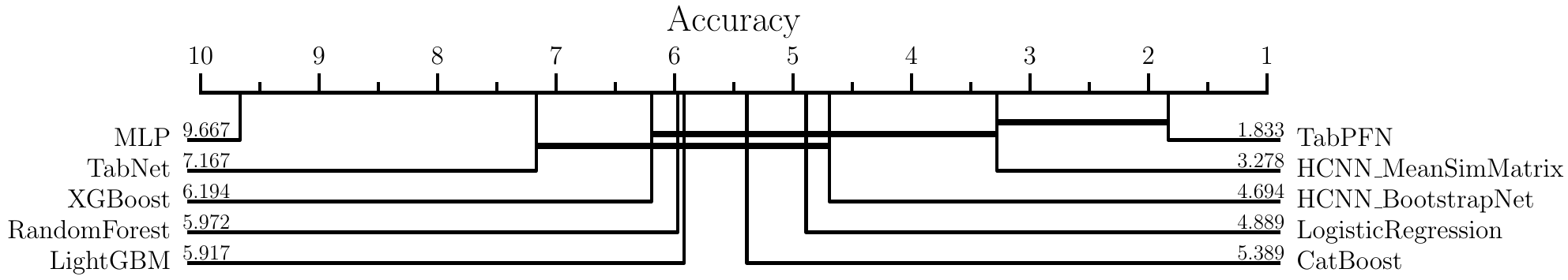}}
    \subfigure[]{\includegraphics[scale=0.43]{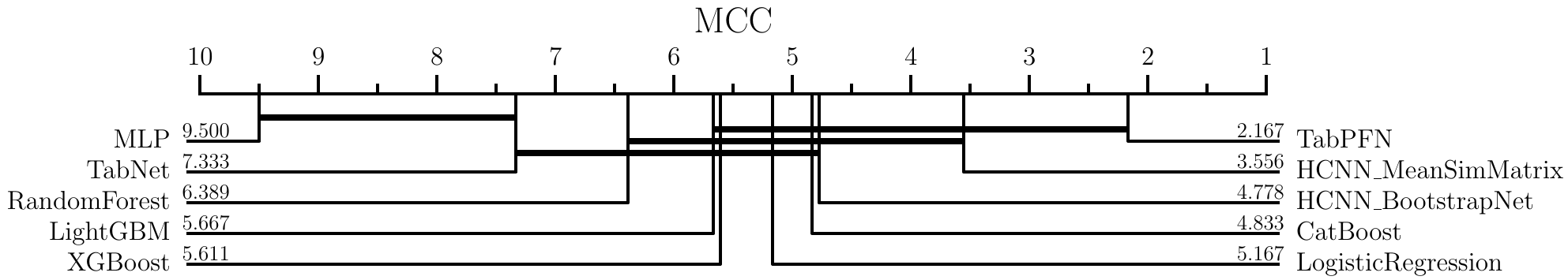}}
    \caption{Critical Difference plots on out-of-sample average ranks with a Wilcoxon significance analysis. In (a) the test is run considering the F1$\_$Score; in (b) the test is run considering the accuracy; in (c) the test is run considering the MCC.}
    \label{fig:Wilcoxon_Test_Small_Datasets_0_1}
    \vspace{-15pt}
\end{figure}

Models' numerical performances for each evaluation metric (see \appendixref{app:Appendix_Dataset_Wise_Scores_Small_Tabular_Data}) enforce all the findings discussed above. It is, however, clear that the differences in performance are very small. To assess their statistical significance, we use the Critical Difference (CD) diagram of the ranks based on the Wilcoxon significance test (with $p$-values below $0.1$), a standard metric for comparing classifiers across multiple datasets \citep{demvsar2006statistical}. The overall empirical comparison of the methods is given in \figureref{fig:Wilcoxon_Test_Small_Datasets_0_1}. We notice that the performance of \verb|HCNN| and \verb|TabPFN| is not statistically different. This finding is coherent across the three evaluation metrics and is particularly relevant because it makes these deep learning architectures the only two which are consistently comparable with the SOTA machine learning ones. These findings legitimate the methodology proposed in the current research work as being itself a SOTA, both in terms of performance and in terms of computational complexity (see \appendixref{app:Appendix_Number_Of_Parameters} for an extended study on the number of parameters). We cannot assert the same for the \verb|TabPFN|, which is among the SOTA models in terms of performance but it is the worst model in terms of computational (and architectural) complexity.

\subsection{Models' scalability to larger tabular numerical classification problems} \label{sec:Models_Scalability}
All the models considered in the current research work are primarily designed to handle small tabular classification problems. According to \citep{hollmann2022tabpfn}, a dataset is defined as “small” if it contains up to $2\,000$ samples and $100$ features. However, in this section, we explore the ability of the models to scale to larger problems. In doing so, we use benchmark datasets characterized, in turn, by a number of samples greater than $2\,000$ or a number of features greater than $100$.

\begin{table}[h]
\vspace{-10pt}
\centering
\caption{Study on models' ability to scale to larger problems. Considered datasets belong to the OpenML benchmark suite “Tabular benchmark numerical classification” \citep{grinsztajn2022tree}. For each of them, we report the OpenML ID, the number of samples, and the number of features. We indicate the success in solving the corresponding tabular classification task with a (\cmark) symbol, while a failure to solve the problem is denoted by an (\xmark) symbol.}
\label{tab:Scalability_Table}
\resizebox{\columnwidth}{!}{%
\begin{tabular}{@{}ccc|cccccccccc@{}}
\toprule
\textbf{OpenML ID} &
  \textbf{\# Samples} &
  \textbf{\# Features} &
  \multicolumn{10}{c}{\textbf{Model}} \\ \midrule
 &
   &
   &
  \textbf{LogisticRegression} &
  \textbf{RandomForest} &
  \textbf{XGBoost} &
  \textbf{LightGBM} &
  \textbf{CatBoost} &
  \textbf{MLP} &
  \textbf{TabNet} &
  \textbf{TabPFN} &
  \textbf{\begin{tabular}[c]{@{}c@{}}HCNN\\ BootstrapNet\end{tabular}} &
  \textbf{\begin{tabular}[c]{@{}c@{}}HCNN\\ MeanSimMatrix\end{tabular}} \\ \midrule
361055 &
  16714 &
  10 &
  \cmark &
  \cmark &
  \cmark &
  \cmark &
  \cmark &
  \cmark &
  \cmark &
  \xmark &
  \cmark &
  \cmark \\
361062 &
  10082 &
  26 &
  \cmark &
  \cmark &
  \cmark &
  \cmark &
  \cmark &
  \cmark &
  \cmark &
  \xmark &
  \cmark &
  \cmark \\
361063 &
  13488 &
  16 &
  \cmark &
  \cmark &
  \cmark &
  \cmark &
  \cmark &
  \cmark &
  \cmark &
  \xmark &
  \cmark &
  \cmark \\
361065 &
  13376 &
  10 &
  \cmark &
  \cmark &
  \cmark &
  \cmark &
  \cmark &
  \cmark &
  \cmark &
  \xmark &
  \cmark &
  \cmark \\
361066 &
  10578 &
  7 &
  \cmark &
  \cmark &
  \cmark &
  \cmark &
  \cmark &
  \cmark &
  \cmark &
  \xmark &
  \cmark &
  \cmark \\
361275 &
  13272 &
  20 &
  \cmark &
  \cmark &
  \cmark &
  \cmark &
  \cmark &
  \cmark &
  \cmark &
  \xmark &
  \cmark &
  \cmark \\
361276 &
  3434 &
  419 &
  \cmark &
  \cmark &
  \cmark &
  \cmark &
  \cmark &
  \cmark &
  \cmark &
  \xmark &
  \cmark &
  \xmark \\
361277 &
  20634 &
  8 &
  \cmark &
  \cmark &
  \cmark &
  \cmark &
  \cmark &
  \cmark &
  \cmark &
  \xmark &
  \cmark &
  \cmark \\
361278 &
  10000 &
  22 &
  \cmark &
  \cmark &
  \cmark &
  \cmark &
  \cmark &
  \cmark &
  \cmark &
  \xmark &
  \cmark &
  \cmark \\ \bottomrule
\end{tabular}%
}
\end{table}

From \tableref{tab:Scalability_Table}, one can observe that the proposed datasets are sufficient in underlining the issues of two models, namely the \verb|TabPFN| model and the \verb|HCNN| model in its \verb|MeanSimMatrix| configuration. In the first case, the model is entirely unable to scale to problems with a larger number of samples and features. This limitation was already pointed out in the original work by \citep{hollmann2022tabpfn} and directly depends on the model's architecture, which strongly leverage the power of attention-based mechanism. Indeed, the runtime and memory usage of the \verb|TabPFN| architecture scales quadratically with the number of inputs (i.e. training samples passed) and the fitted model cannot work with datasets with a number of features $> 100$. In the case of \verb|HCNN MeanSimMatrix|, instead, the proposed architecture demonstrates a limit in handling problems characterised by a large number of features (but not samples). Also in this case, the reason of the failure should be searched in the model's architectural design choices. Indeed, this architecture is characterised by a strong linear relationship between the number of features and the number of parameters (see \appendixref{app:Appendix_Number_Of_Parameters}), meaning that when the first parameter is large, convolving across all representatives of each simplicial complex family becomes computationally demanding. A solution to this problem can be found in employing the \verb|BootstrapNet| configuration, which disrupts the linear relationship discussed earlier, resulting in a significant reduction in the number of parameters when dealing with a large number of features. While this approach demonstrates considerable efficacy, it remains reliant on a threshold parameter (see \sectionref{sec:Information_Filtering_Networks}), suggesting the need for more advanced and parameter-free alternatives. For the seek of completeness, in \appendixref{app:Appendix_Results_Scalability_Datasets} we partially repeat the analyses presented in \sectionref{sec:Small_tabular_classification_results} on the newly introduced datasets. Because of the fragmentation caused by the increased size, we report only the dataset-dependent analyses, excluding cross-datasets ones.

\section{Conclusion} \label{sec:Conclusion}
In this paper, we introduce the Homological Convolutional Neural Network (\verb|HCNN|), a novel deep learning architecture that revisits the simpler Homological Neural Network (\verb|HNN|) to gain abstraction, representation power, robustness, and scalability. The proposed architecture is data-centric and arises from a graph-based higher-order representation of dependency structures among multivariate input features. Compared to \verb|HNN|, our model demonstrates a higher level of abstraction. Indeed, we have higher flexibility in choosing the initial network's representation, as we can choose from the universe simplicial complexes and we are not restricted to specific sub-families. Looking at geometrical structures at different granularity levels, we propose a clear-cut way to leverage the power of convolution on sparse data representations. This allows to fully absorb the representation power of \verb|HNN| in the very first layer of \verb|HCNN|, leaving room for additional data transformations at deeper levels of the architecture. Specifically, in the current research work, we build the \verb|HCNN| using a class of information filtering networks (i.e. the TMFG) that uses squared correlation coefficients to maximize the likelihood of the underlying system. We propose two alternative architectural solutions: (i) the \verb|MeanSimMatrix| configuration; and (ii) the \verb|BootstrapNet| configuration. Both of them leverage the power of bootstrapping to gain robustness toward data noise and the intrinsic complexity of interactions among the underlying system's variables. We test these two modeling solutions on a set of tabular numerical classification problems (i.e. one of the most challenging tasks for deep learning models and the one where \verb|HNN| demonstrates the poorest performances). We compare \verb|HCNN| with different machine- and deep learning architectures, always teeing SOTA performances and demonstrating superior robustness to data unbalances. Specifically, we demonstrate that \verb|HCNN| is able to compete with the latest transformer-based architectures (e.g. \verb|TabPFN|) by using a considerably lower and easily controllable number of parameters (especially in the \verb|BootstrapNet| configuration), guaranteeing a higher level of explicability in the neural network's building process and having a comparable running time without the need for pre-training. We finally propose a study on models' scalability to large datasets. We underline the fragility of transformer-based models and we demonstrate that \verb|HCNN| in its \verb|MeanSimMatrix| configuration is unable to manage datasets characterized by a large number of input features. In contrast, we show that the design choices adopted for the \verb|BootstrapNet| configuration offers a parametric solution to the problem. Despite significant advances introduced by \verb|HCNN|s, this class of neural networks remains in an embryonic phase. Further studies on underlying network representations should propose alternative similarity metrics that replace the squared correlation coefficients for mixed data-types (i.e. categorical and numerical or categorical only data-types), and further work is finally required to better understand low-level interactions captured by the proposed neural network model. This final point would certainly lead to a class of non-parametric, parsimonious \verb|HCNN|s. 

\newpage

\bibliography{pmlr-sample}

\newpage

\appendix

\section{}\label{app:Appendix_Small_Tabular_Data}

Tables \ref{tab:datasets_description_table} and \ref{tab:scalability_datasets_description} report an overview of the main characteristics of the two suites of benchmark datasets used in the current research work. In both cases the open-access of data is guaranteed by OpenML \citep{OpenML2021}. Training/validation/test split is not provided. For all the datasets, the $50\%$ of the raw dataset is used as a training set, the $25\%$ as validation set, and the remaining $25\%$ as a test set. To prove the statistical significance of results presented in the current research work, all the analyses are repeated on $10$ different combinations of training/validation/test splits. The reproducibility of results is guaranteed by a rigorous usage of seeds (i.e. $[12,\,\,190,\,\,903,\,\,7\,687,\,\,8\,279,\,\,9\,433,\,\,12\,555,\,\,22\,443,\,\,67\,822, 9\,822\,127]$).

\begin{table}[H]
    \vspace{-15pt}
    \centering
    \caption{Datasets used for models' evaluation. These include 18 numerical tabular datasets from the OpenML-CC18 benchmark suite with at most $2\,000$ samples, $100$ features and $10$ classes. For each dataset we report the name, the number of input features, the number of samples before training/validation/test split, the number of classes, the number of samples by class and, finally, the corresponding ID number in the OpenML benchmark suite.}
    \label{tab:datasets_description_table}
    \resizebox{\columnwidth}{!}{%
    \begin{tabular}{@{}cccccc@{}}
    \toprule
    \textbf{Dataset Name} &
      \textbf{\# Features} &
      \textbf{\# Samples} &
      \textbf{\# Classes} &
      \textbf{\begin{tabular}[c]{@{}c@{}}Samples by \\ Class\end{tabular}} &
      \textbf{OpenML ID} \\ \midrule
    balance-scale                    & 4  & 625  & 3  & 49/288/288                                                                         & 11    \\ \midrule
    mfeat-fourier                    & 76 & 2,000 & 10 & \begin{tabular}[c]{@{}c@{}}200/200/200/200/200/\\ 200/200/200/200/200\end{tabular} & 14    \\ \midrule
    breast-w                         & 9  & 683  & 2  & 444/239                                                                            & 15    \\ \midrule
    mfeat-karhunen                   & 64 & 2,000 & 10 & \begin{tabular}[c]{@{}c@{}}200/200/200/200/200/\\ 200/200/200/200/200\end{tabular} & 16    \\ \midrule
    mfeat-morphological &
      6 &
      2,000 &
      10 &
      \begin{tabular}[c]{@{}c@{}}200/200/200/200/200/\\ 200/200/200/200/200\end{tabular} &
      18 \\ \midrule
    mfeat-zernike                    & 47 & 2,000 & 10 & \begin{tabular}[c]{@{}c@{}}200/200/200/200/200/\\ 200/200/200/200/200\end{tabular} & 22    \\ \midrule
    diabetes                         & 8  & 768  & 2  & 500/268                                                                            & 37    \\ \midrule
    vehicle                          & 18 & 846  & 4  & 218/212/217/199                                                                    & 54    \\ \midrule
    analcatdata\_authorship          & 70 & 841  & 4  & 317/296/55/173                                                                     & 458   \\ \midrule
    pc4                              & 37 & 1458 & 2  & 1280/178                                                                           & 1049  \\ \midrule
    kc2                              & 21 & 522  & 2  & 415/107                                                                            & 1063  \\ \midrule
    pc1                              & 21 & 1109 & 2  & 1032/77                                                                            & 1068  \\ \midrule
    banknote-authentication          & 4  & 1372 & 2  & 762/610                                                                            & 1462  \\ \midrule
    blood-transfusion-service-center & 4  & 748  & 2  & 570/178                                                                            & 1464  \\ \midrule
    qsar-biodeg                      & 41 & 1055 & 2  & 699/356                                                                            & 1494  \\ \midrule
    wdbc                             & 30 & 569  & 2  & 357/212                                                                            & 1510  \\ \midrule
    steel-plates-fault               & 27 & 1941 & 7  & \begin{tabular}[c]{@{}c@{}}402/55/391/\\ 673/158/72/190\end{tabular}               & 40982 \\ \midrule
    climate-model-simulation-crashes & 18 & 540  & 2  & 46/494                                                                             & 40994 \\ \bottomrule
    \end{tabular}%
    }
\end{table}

The first set of data belongs to the “OpenML-CC18” benchmark suite \citep{hollmann2022tabpfn} and is used to compare models' performance in solving numerical classification problems, while the second set of data belongs to the “OpenML tabular benchmark numerical classification” suite \citep{grinsztajn2022tree} and is used to test the models' ability to scale to larger problems of the same type. Looking at \tableref{tab:datasets_description_table}, we notice that the average number of features is $28.5$ with a standard deviation equal to $22.2$. The dataset with the lowest number of features is “balance-scale” (i.e. $4$ features) with OpenML ID $11$. The dataset with the largest number of features is “mfeat-fourier” (i.e. $76$ features) with OpenML ID $14$. All the features in all the datasets are numerical only and no missing values are detected. The average number of samples is $1191.6$ with a standard deviation equal to $557.3$. The dataset with the lowest number of samples is “kc2” (i.e. $522$ samples) with OpenML ID $1063$. The dataset with the largest number of samples is “mfeat-morphological” (i.e. $2\,000$ samples) with OpenML ID $18$. $11$ of the considered datasets are binary, while $8$ are multi-class.

\begin{table}[H]
\vspace{-10pt}
\centering
\caption{Datasets used to test models' scalability. These include $9$ numerical tabular datasets from the OpenML “Tabular benchmark numerical classification” suite \citep{grinsztajn2022tree}. All these datasets violate at least one of the selection criteria in \citep{hollmann2022tabpfn} (i.e. they are characterised by a number of samples $> 2\,000$ or they are characterised by a number of features $> 100$). For each dataset we report the name, the number of input features, the number of samples before training/validation/test split, the number of classes, the number of samples by class and, finally, the corresponding ID number in the OpenML benchmark suite.}
\label{tab:scalability_datasets_description}
\resizebox{\columnwidth}{!}{%
\begin{tabular}{@{}cccccc@{}}
\toprule
\textbf{Dataset Name} &
  \textbf{\# Features} &
  \textbf{\# Samples} &
  \textbf{\# Classes} &
  \textbf{\begin{tabular}[c]{@{}c@{}}Samples by \\ Class\end{tabular}} &
  \textbf{OpenML ID} \\ \midrule
credit                         & 10  & 16714 & 2 & 8357/8357   & 361055 \\ \midrule
pol                            & 26  & 10082 & 2 & 5041/5041   & 361062 \\ \midrule
house\_16H                     & 16  & 13488 & 2 & 6744/6744   & 361063 \\ \midrule
MagicTelescope                 & 10  & 13376 & 2 & 6688/6688   & 361065 \\ \midrule
bank-marketing                 & 7   & 10578 & 2 & 5289/5289   & 361066 \\ \midrule
default-of-credit-card-clients & 20  & 13272 & 2 & 6636/6636   & 361275 \\ \midrule
Bioresponse                    & 419 & 3434  & 2 & 1717/1717   & 361276 \\ \midrule
california                     & 8   & 20634 & 2 & 10317/10317 & 361277 \\ \midrule
heloc                          & 22  & 10000 & 2 & 5000/5000   & 361278 \\ \bottomrule
\end{tabular}%
}
\end{table}

Looking at \tableref{tab:scalability_datasets_description}, we notice that the average number of features is $59.8$ with a standard deviation equal to $127.2$. The dataset with the lowest number of features is “bank-marketing” (i.e. $7$ features) with OpenML ID $361066$. The dataset with the largest number of features is “Bioresponse” (i.e. $419$ features) with OpenML ID $361276$. All the features in all the datasets are numerical only and no missing values are detected. The average number of samples is $12397.6$ with a standard deviation equal to $4523.4$. The dataset with the lowest number of samples is “Bioresponse” (i.e. $3434$ samples) with OpenML ID $361276$. The dataset with the largest number of samples is “california” (i.e. $20634$ samples) with OpenML ID $361277$. All the considered datasets are binary.

\newpage

\section{}\label{app:Appendix_TMFG_Algo}

\begin{algorithm2e}
    \caption{TMFG built on the similarity matrix $\hat{\textbf{C}}$ to maximize global properties of the system under analysis (e.g., likelihood).}\label{alg:TMFG_Algo}
    \KwData{Similarity matrix $\hat{\textbf{C}} \in \mathbb{R}^{n, n}$ from a set of observations $\{x_{1, 1}, \dots, x_{s, 1}\}, \{x_{1, 2}, \dots, x_{s, 2}\} \dots \{x_{1, n}, \dots, x_{s, n}\}$.}
    \KwResult{Sparse adjacency matrix \textbf{\textit{A}} describing the TMFG.}
    
    \SetAlgoNlRelativeSize{0}
    \SetNlSty{textbf}{(}{)}
    \SetAlgoNlRelativeSize{-1}
    
    \DontPrintSemicolon

    \SetKwProg{Fn}{Function}{:}{}
    
    \Fn{MaximumGain($\hat{\textbf{C}}$, $\mathcal{V}$, $t$)}{
        Initialize a vector of zeros $g \in \mathbb{R}^{1 \times n}$\;
        \For{$j \in t$}{
            \For{$v \notin \mathcal{V}$}{
                $\hat{\textbf{C}}_{v,j} = 0$\;
            }
            $g = g \oplus \hat{\textbf{C}}_{v,j}$\;
        }
        \Return $\max{\{g\}}$\;
    }

    Initialize four empty sets: $\mathcal{C}$ (cliques), $\mathcal{T}$ (triangles), $\mathcal{S}$ (separators), and $\mathcal{V}$ (vertices)\;
    Initialize an adjacency matrix $\textbf{\textit{A}} \in \mathbb{R}^{n, n}$ with all zeros\;
    $\mathcal{C}_1 \leftarrow$ tetrahedron, $\{v_1, v_2, v_3, v_4\}$, obtained by choosing the $4$ entries of $\hat{\textbf{C}}$ maximizing the similarity among features\;
    $\mathcal{T} \leftarrow$ the four triangular faces in $\mathcal{C}_1$: $\{v_1, v_2, v_3\}$, $\{v_1, v_2, v_4\}$, $\{v_1, v_3, v_4\}$, $\{v_2, v_3, v_4\}$\;
    $\mathcal{V} \leftarrow$ Assign to $\mathcal{V}$ the remaining $n-4$ vertices not in $\mathcal{C}_1$\;
    
    \While{$\mathcal{V}$ is not empty}{
        Find the combination of $\{v_a, v_b, v_c\} \in \mathcal{T}$ (i.e. $t$) and $v_d \in \mathcal{V}$ which maximizes \texttt{MaximumGain($\hat{\textbf{C}}$, $\mathcal{V}$, $t$)}\;
        /* $\{v_a, v_b, v_c, v_d\}$ is a new 4-clique $\mathcal{C}$, $\{v_a, v_b, v_c\}$ becomes a separator $\mathcal{S}$, three new triangular faces, $\{v_a, v_b, v_d\}$, $\{v_a, v_c, v_d\}$, and $\{v_b, v_c, v_d\}$ are created */\;
        Remove $v_d$ from $\mathcal{V}$\;
        Remove $\{v_a, v_b, v_c\}$ from $\mathcal{T}$\;
        Add $\{v_a, v_b, v_d\}$, $\{v_a, v_c, v_d\}$, and $\{v_b, v_c, v_d\}$ to $\mathcal{T}$\;
    }
    
    \For{each pair of nodes $i, j$ in $\mathcal{C}$}{
        Set $\textbf{\textit{A}}_{i,j} = 1$\;
    }
    
    \Return $\textbf{\textit{A}}$\;
\end{algorithm2e}

\newpage

\section{}\label{app:Appendix_HNN_HCNN}

\begin{figure}[H]\label{fig:HNN_HCNN_Building_Pipeline}
    \vspace{-5pt}
    \centering
    \subfigure[]{\includegraphics[scale=0.15]{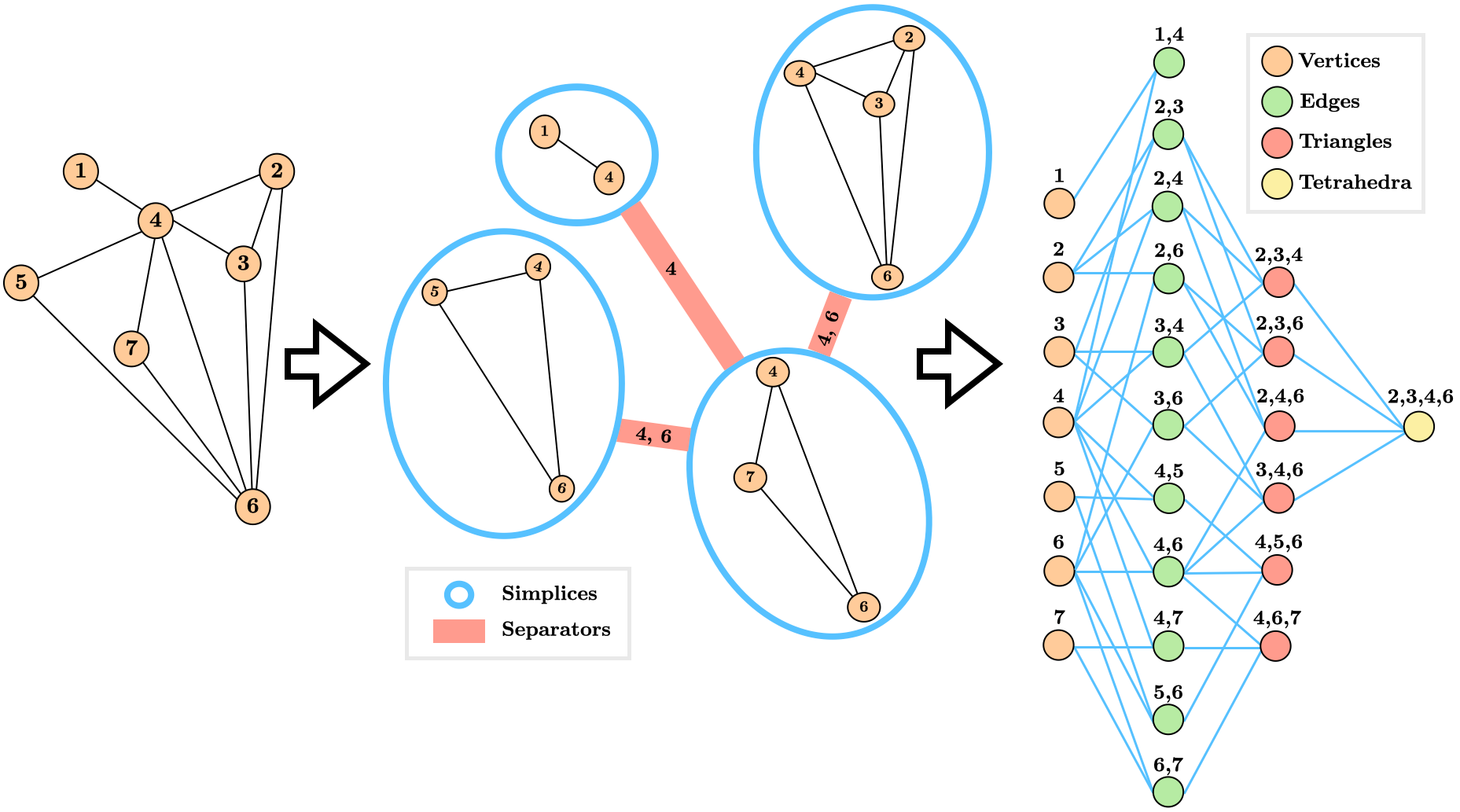}}
    \subfigure[]{\includegraphics[scale=0.15]{Images/HCNN.png}}
    \caption{Pictorial representations of an HNN and HCNN architectures. In Figure (a), from left to right, (i) we start from a chordal graph representing the dependency structures of features in the underlying system; (ii) we re-arrange the network's representation to highlight the underlying simplicial complex structures (i.e. edges, triangles, tetrahedra); and (iii) we finally report a layered representation, which explicitly takes into account higher order sub-structures and their interconnections, and can be easily converted into a computational unit (i.e. a sparse MLP). In Figure (b), from left to right, (i) we start from a chordal graph representing the dependency structures of features in the underlying system; (ii) we isolate the maximal cliques corresponding to $1$-, $2$- and $3$-dimensional simplices (i.e. edges, triangles, tetrahedra) and we group them into $1$D vectors containing features' realizations; (iii) we compute a $1^{st}$-level convolution to extract simplicial-wise non-linear relationships; (iv) we compute a $2^{nd}$-level convolution, which operates on the output of the previous level of convolution across all the representatives of each simplicial family extracting a class of non-trivial homological insights; (v) we finally apply a linear map from the $2^{nd}$-level convolution to the output, extracting model's predictions.}
    \vspace{-15pt}
\end{figure}

\section{}\label{app:Appendix_Hyperparameters}

\begin{table}[H]
\centering
\caption{Hyper-parameters' search space for all the classifiers considered in the current paper. When possible, search space is inherited from \citep{shwartz2018representation} and \citep{hollmann2022tabpfn}.}
\label{tab:models_hyperparameter_space}
\scalebox{0.75}{%
\begin{tabular}{ccccc}
\hline
\textbf{Model} &
  \textbf{Name} &
  \textbf{Type} &
  \textbf{Value} &
  \textbf{Skip} \\ \hline
LogisticRegression &
  \begin{tabular}[c]{@{}c@{}}penalty\\ max\_iter\end{tabular} &
  \begin{tabular}[c]{@{}c@{}}cat\\ int\end{tabular} &
  \begin{tabular}[c]{@{}c@{}}(l1, l2, elasticnet)\\ ($100$, $500$, $1000$)\end{tabular} &
  \begin{tabular}[c]{@{}c@{}}-\\ -\end{tabular} \\ \hline
RandomForest &
  \begin{tabular}[c]{@{}c@{}}n\_estimators\\ max\_depth\\ min\_samples\_leaf\\ min\_samples\_split\end{tabular} &
  \begin{tabular}[c]{@{}c@{}}int\\ int\\ int\\ int\end{tabular} &
  \begin{tabular}[c]{@{}c@{}}{[}$100$, $4000${]}\\ {[}$10$, $50${]}\\ {[}$2$, $10${]}\\ {[}$2$, $10${]}\end{tabular} &
  \begin{tabular}[c]{@{}c@{}}$200$\\ $10$\\ $2$\\ $2$\end{tabular} \\ \hline
XGBoost &
  \begin{tabular}[c]{@{}c@{}}n\_estimators\\ max\_depth\\ learning\_rate\\ subsample\\ colsample\_bytree\\ colsample\_bylevel\\ alpha\\ lambda\\ gamma\end{tabular} &
  \begin{tabular}[c]{@{}c@{}}int\\ int\\ float\\ float\\ float\\ float\\ float\\ float\\ float\end{tabular} &
  \begin{tabular}[c]{@{}c@{}}{[}$100$, $4000${]}\\ {[}$1$, $10${]}\\ {[}$e^{-4}$, $1${]}\\ {[}$0.2$, $1${]}\\ {[}$0.2$, $1${]}\\ {[}$0.2$, $1${]}\\ {[}$e^{-4}$, $e^2${]}\\ {[}$e^{-4}$, $e^2${]}\\ {[}$e^{-4}$, $e^2${]}\end{tabular} &
  \begin{tabular}[c]{@{}c@{}}$200$\\ $3$\\ -\\ -\\ -\\ -\\ -\\ -\\ -\end{tabular} \\ \hline
CatBoost &
  \begin{tabular}[c]{@{}c@{}}n\_estimators\\ max\_depth\\ learning\_rate\\ random\_strength\\ l2\_leaf\_reg\\ bagging\_temperature\\ leaf\_estimation\_iterations\end{tabular} &
  \begin{tabular}[c]{@{}c@{}}int\\ int\\ float\\ int\\ int\\ float\\ int\end{tabular} &
  \begin{tabular}[c]{@{}c@{}}{[}$100$, $4000${]}\\ {[}$1$, $10${]}\\ {[}$e^{-4}$, $1${]}\\ {[}$1$, $10${]}\\ {[}$1$, $10${]}\\ {[}$0$, $1${]}\\ {[}$1$, $10${]}\end{tabular} &
  \begin{tabular}[c]{@{}c@{}}$300$\\ $3$\\ -\\ $3$\\ $3$\\ -\\ $3$\end{tabular} \\ \hline
LightGBM &
  \begin{tabular}[c]{@{}c@{}}n\_estimators\\ max\_depth\\ learning\_rate\\ num\_leaves\\ reg\_alpha\\ \\ reg\_lambda\\ \\ subsample\end{tabular} &
  \begin{tabular}[c]{@{}c@{}}int\\ int\\ float\\ int\\ float\\ \\ float\\ \\ float\end{tabular} &
  \begin{tabular}[c]{@{}c@{}}{[}$100$, $4000${]}\\ {[}$1$, $10${]}\\ {[}$e^{-4}$, $1${]}\\ {[}$5$, $50${]}\\ ($0$, $0.01$, $1$,  $2$,\\ $5$, $7$, $10$, $50$, $100$)\\ ($0$, $0.01$, $1$, $5$, $10$, \\ $20$, $50$, $100$)\\ {[}$0.2$, $0.8${]}\end{tabular} &
  \begin{tabular}[c]{@{}c@{}}$300$\\ $3$\\ -\\ $5$\\ -\\ \\ -\\ \\ -\end{tabular} \\ \hline
MLP &
  \begin{tabular}[c]{@{}c@{}}hidded\_layer\_sizes\\ alpha\\ max\_iter\\ learning\_rate\end{tabular} &
  \begin{tabular}[c]{@{}c@{}}int\\ float\\ int\\ cat\end{tabular} &
  \begin{tabular}[c]{@{}c@{}}($10$, $50$, $100$, $150$, $200$)\\ ($0.1$, $0.01$, $0.001$, $0.0001$)\\ ($100$, $500$, $1000$)\\ (constant, invscaling, adaptive)\end{tabular} &
  \begin{tabular}[c]{@{}c@{}}-\\ -\\ -\\ -\end{tabular} \\ \hline
TabNet &
  \begin{tabular}[c]{@{}c@{}}learning\_rate\\ n\_steps\\ relaxation\_factor\end{tabular} &
  \begin{tabular}[c]{@{}c@{}}float\\ int\\ float\end{tabular} &
  \begin{tabular}[c]{@{}c@{}}{[}$e^{-4}$, $1${]}\\ {[}$1$,  $8${]}\\ {[}$0.3$, $2${]}\end{tabular} &
  \begin{tabular}[c]{@{}c@{}}-\\ -\\ -\end{tabular} \\ \hline
TabPFN &
  n\_ensemble\_configurations &
  int &
  {[}$8$, $128${]} &
  $8$ \\ \hline
\begin{tabular}[c]{@{}c@{}}HCNN\\ BootstrapNet\end{tabular} &
  \begin{tabular}[c]{@{}c@{}}n\_filters\_l1\\ n\_filters\_l2\\ tmfg\_iterations\\ tmfg\_confidence\\ tmfg\_similarity\end{tabular} &
  \begin{tabular}[c]{@{}c@{}}int\\ int\\ int\\ float\\ cat\end{tabular} &
  \begin{tabular}[c]{@{}c@{}}{[}$4$, $16${]}\\ {[}$32$, $64${]}\\ {[}$100$, $1000${]}\\ ($0.90$, $0.95$, $0.99$)\\ (pearson, spearman)\end{tabular} &
  \begin{tabular}[c]{@{}c@{}}$4$\\ $4$\\ $300$\\ -\\ -\end{tabular} \\ \hline
\begin{tabular}[c]{@{}c@{}}HCNN\\ MeanSimMatrix\end{tabular} &
  \begin{tabular}[c]{@{}c@{}}n\_filters\_l1\\ n\_filters\_l2\\ tmfg\_iterations\\ tmfg\_similarity\end{tabular} &
  \begin{tabular}[c]{@{}c@{}}int\\ int\\ int\\ cat\end{tabular} &
  \begin{tabular}[c]{@{}c@{}}{[}$4$, $16${]}\\ {[}$32$, $64${]}\\ {[}$100$, $1000${]}\\ (pearson, spearman)\end{tabular} &
  \begin{tabular}[c]{@{}c@{}}$4$\\ $4$\\ $300$\\ -\end{tabular} \\ \hline
\end{tabular}%
}
\end{table}

\newpage

\section{}\label{app:Appendix_Radar_Plots}
\begin{figure}[h]\label{fig:Radar_Plot_Small_Tabular_Classification_Problems}
    \centering
    \subfigure[]{{\includegraphics[width=0.48\textwidth]{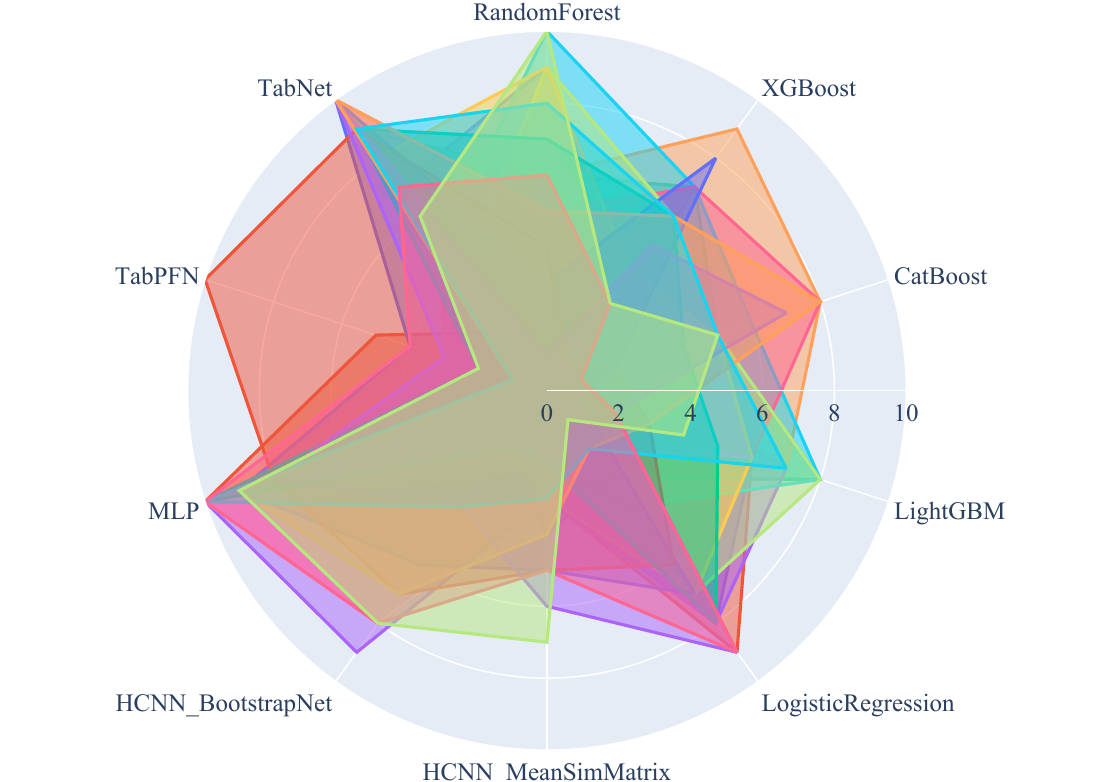}}\label{fig:RadarPlot_Classification_Small_F1_Score}}
    \subfigure[]{{\includegraphics[width=0.5\textwidth]{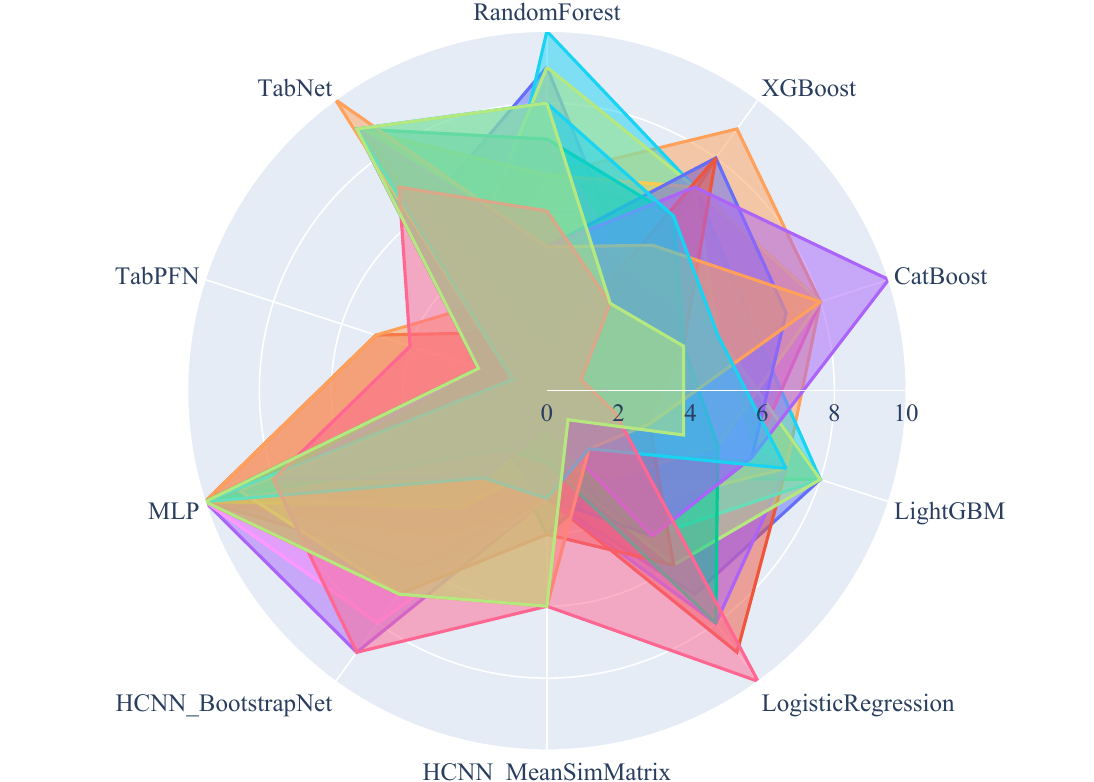}}\label{fig:RadarPlot_Classification_Small_Accuracy_Score}}
    \subfigure[]{{\includegraphics[scale=0.5]{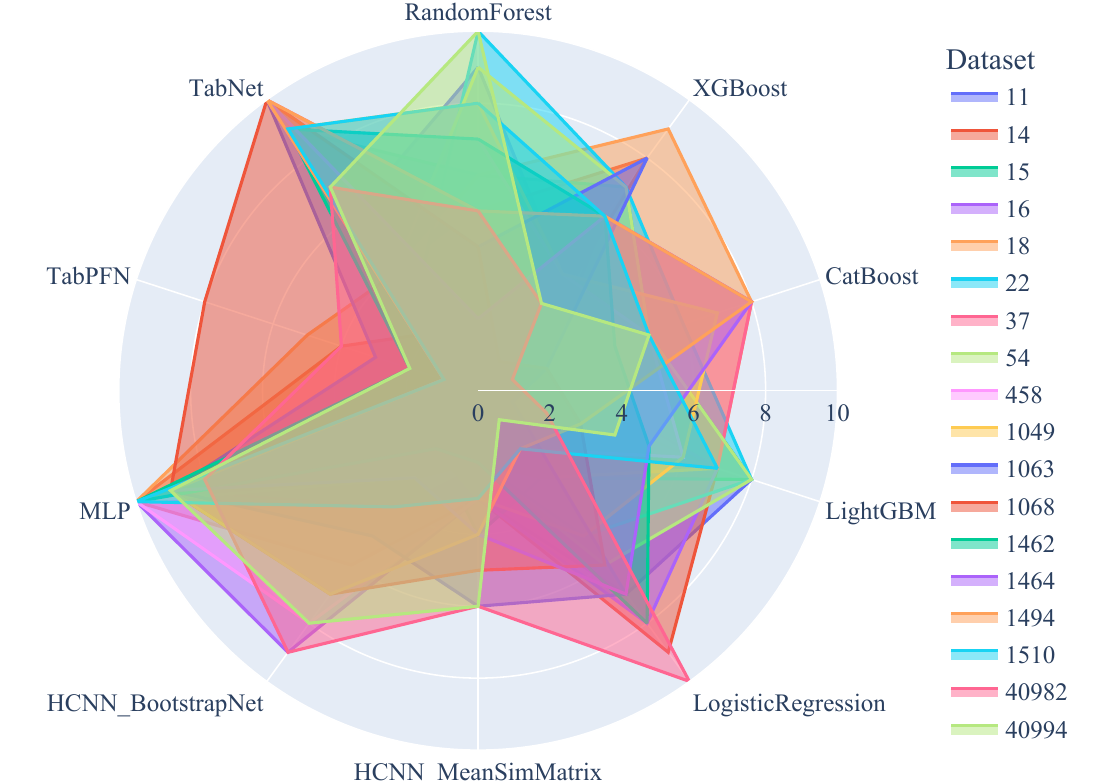}}\label{fig:RadarPlot_Classification_Small_MCC_Score}}
    \caption{Out-of-sample model- and dataset-dependent average ranking considering (a) F1$\_$Score, (b) Accuracy, and (c) MCC evaluation metric. This representation allows to clearly assess the higher robustness of HCNN model to datasets' unbalance over all its deep learning and machine learning alternatives.}
    \label{fig:Extensive_RadarPlots}
\end{figure}

All the findings related to \tableref{tab:Small_Tabular_Classification_Compact_Results} and discussed in \sectionref{sec:Small_tabular_classification_results} can be graphically visualized in Figure \ref{fig:Extensive_RadarPlots}. Specifically, the highest robustness of \verb|HCNN| model in its \verb|MeanSimMatrix| configuration compared to \verb|TabPFN| model can be observed in Figure \ref{fig:RadarPlot_Classification_Small_F1_Score} and Figure \ref{fig:RadarPlot_Classification_Small_MCC_Score}. They report the ranking position of each model on each dataset belonging to the "OpenML-CC18" benchmark suite using the F1\_Score and the MCC as performance metrics respectively. In the first case, we notice that the worst ranking position by \verb|HCNN| is $7$ and is reached when dealing with dataset “climate-model-simulation-crashes” (OpenML ID $40994$), while the one occupied by \verb|TabPFN| is $10$ with dataset “pc\_1” (OpenML ID $1068$). In the second case, we notice that the worst performance by \verb|HCNN| has ranking $6$ and is reached when dealing with datasets “mfeat-karhunen” (OpenML ID $16$), “steel-plates-fault” (OpenML ID $40982$) and “climate-model-simulation-crashes” (OpenML ID $40994$), while the one occupied by \verb|TabPFN| is $8$ with dataset “pc\_1” (OpenML ID $1068$). Except for “mfeat-karhunen” dataset (OpenML ID $16$), all the datasets listed before are strongly unbalanced.

\newpage

\section{}\label{app:Appendix_Dataset_Wise_Scores_Small_Tabular_Data}

\begin{table}[H]
\vspace{-15pt}
\centering
\caption{For each considered model, we report the average F1\_score and the corresponding standard deviation (across $10$ different seeds) on the $18$ benchmark datasets from the “OpenML-CC-18” benchmark suite.}
\label{tab:average_std_f1_score_small_tabular_data}
\scalebox{0.43}{%
\begin{tabular}{@{}c|cccccccccc@{}}
\toprule
\textbf{Dataset ID} &
  \multicolumn{10}{c}{\textbf{Model}} \\ \midrule
\textbf{} &
  \textbf{LogisticRegression} &
  \textbf{RandomForest} &
  \textbf{XGBoost} &
  \textbf{LightGBM} &
  \textbf{CatBoost} &
  \textbf{MLP} &
  \textbf{TabNet} &
  \textbf{TabPFN} &
  \textbf{\begin{tabular}[c]{@{}c@{}}HCNN\\ BootstrapNet\end{tabular}} &
  \textbf{\begin{tabular}[c]{@{}c@{}}HCNN\\ MeanSimMatrix\end{tabular}} \\ \midrule
\textbf{11} &
  0.62$\pm$0.03 &
  0.60$\pm$0.01 &
  0.93$\pm$0.03 &
  0.71$\pm$0.07 &
  0.73$\pm$0.04 &
  0.59$\pm$0.03 &
  0.65$\pm$0.05 &
  0.97$\pm$0.02 &
  0.95$\pm$0.04 &
  0.94$\pm$0.04 \\
\textbf{14} &
  0.79$\pm$0.01 &
  0.82$\pm$0.02 &
  0.81$\pm$0.02 &
  0.81$\pm$0.02 &
  0.82$\pm$0.01 &
  0.77$\pm$0.03 &
  0.82$\pm$0.01 &
  0.82$\pm$0.01 &
  0.81$\pm$0.01 &
  0.82$\pm$0.02 \\
\textbf{15} &
  0.97$\pm$0.01 &
  0.96$\pm$0.01 &
  0.96$\pm$0.01 &
  0.96$\pm$0.01 &
  0.96$\pm$0.01 &
  0.95$\pm$0.02 &
  0.95$\pm$0.02 &
  0.97$\pm$0.01 &
  0.97$\pm$0.01 &
  0.96$\pm$0.01 \\
\textbf{16} &
  0.95$\pm$0.01 &
  0.96$\pm$0.01 &
  0.95$\pm$0.00 &
  0.95$\pm$0.01 &
  0.96$\pm$0.01 &
  0.94$\pm$0.01 &
  0.96$\pm$0.01 &
  0.97$\pm$0.01 &
  0.94$\pm$0.01 &
  0.96$\pm$0.01 \\
\textbf{18} &
  0.73$\pm$0.02 &
  0.72$\pm$0.02 &
  0.71$\pm$0.03 &
  0.72$\pm$0.01 &
  0.72$\pm$0.02 &
  0.68$\pm$0.02 &
  0.72$\pm$0.02 &
  0.73$\pm$0.02 &
  0.73$\pm$0.02 &
  0.73$\pm$0.02 \\
\textbf{22} &
  0.82$\pm$0.01 &
  0.77$\pm$0.02 &
  0.79$\pm$0.01 &
  0.78$\pm$0.01 &
  0.79$\pm$0.02 &
  0.77$\pm$0.03 &
  0.80$\pm$0.03 &
  0.82$\pm$0.02 &
  0.82$\pm$0.02 &
  0.82$\pm$0.01 \\
\textbf{37} &
  0.73$\pm$0.04 &
  0.72$\pm$0.03 &
  0.72$\pm$0.04 &
  0.72$\pm$0.03 &
  0.71$\pm$0.03 &
  0.70$\pm$0.04 &
  0.68$\pm$0.08 &
  0.73$\pm$0.03 &
  0.72$\pm$0.04 &
  0.72$\pm$0.02 \\
\textbf{54} &
  0.76$\pm$0.03 &
  0.73$\pm$0.03 &
  0.77$\pm$0.04 &
  0.76$\pm$0.04 &
  0.77$\pm$0.04 &
  0.67$\pm$0.06 &
  0.80$\pm$0.04 &
  0.84$\pm$0.03 &
  0.82$\pm$0.03 &
  0.83$\pm$0.03 \\
\textbf{458} &
  1.00$\pm$0.00 &
  0.98$\pm$0.01 &
  0.99$\pm$0.01 &
  0.98$\pm$0.01 &
  0.99$\pm$0.01 &
  0.95$\pm$0.04 &
  0.96$\pm$0.03 &
  1.00$\pm$0.00 &
  0.97$\pm$0.04 &
  0.99$\pm$0.01 \\
\textbf{1049} &
  0.73$\pm$0.03 &
  0.69$\pm$0.04 &
  0.75$\pm$0.03 &
  0.74$\pm$0.05 &
  0.74$\pm$0.04 &
  0.63$\pm$0.08 &
  0.70$\pm$0.10 &
  0.77$\pm$0.04 &
  0.75$\pm$0.03 &
  0.75$\pm$0.04 \\
\textbf{1063} &
  0.68$\pm$0.05 &
  0.70$\pm$0.04 &
  0.64$\pm$0.11 &
  0.71$\pm$0.05 &
  0.70$\pm$0.03 &
  0.62$\pm$0.09 &
  0.58$\pm$0.09 &
  0.69$\pm$0.06 &
  0.68$\pm$0.06 &
  0.68$\pm$0.05 \\
\textbf{1068} &
  0.57$\pm$0.03 &
  0.60$\pm$0.06 &
  0.64$\pm$0.04 &
  0.63$\pm$0.05 &
  0.64$\pm$0.06 &
  0.52$\pm$0.05 &
  0.52$\pm$0.05 &
  0.52$\pm$0.05 &
  0.54$\pm$0.05 &
  0.58$\pm$0.03 \\
\textbf{1462} &
  0.98$\pm$0.01 &
  0.99$\pm$0.00 &
  0.99$\pm$0.00 &
  1.00$\pm$0.00 &
  1.00$\pm$0.00 &
  0.94$\pm$0.05 &
  0.97$\pm$0.04 &
  1.00$\pm$0.00 &
  1.00$\pm$0.00 &
  1.00$\pm$0.00 \\
\textbf{1464} &
  0.56$\pm$0.06 &
  0.64$\pm$0.05 &
  0.62$\pm$0.04 &
  0.64$\pm$0.03 &
  0.60$\pm$0.05 &
  0.56$\pm$0.08 &
  0.54$\pm$0.10 &
  0.63$\pm$0.06 &
  0.62$\pm$0.05 &
  0.61$\pm$0.06 \\
\textbf{1494} &
  0.85$\pm$0.02 &
  0.83$\pm$0.02 &
  0.83$\pm$0.02 &
  0.84$\pm$0.02 &
  0.82$\pm$0.02 &
  0.81$\pm$0.04 &
  0.78$\pm$0.13 &
  0.86$\pm$0.01 &
  0.83$\pm$0.02 &
  0.84$\pm$0.02 \\
\textbf{1510} &
  0.97$\pm$0.02 &
  0.94$\pm$0.02 &
  0.95$\pm$0.02 &
  0.94$\pm$0.02 &
  0.95$\pm$0.01 &
  0.93$\pm$0.02 &
  0.93$\pm$0.02 &
  0.97$\pm$0.01 &
  0.96$\pm$0.02 &
  0.96$\pm$0.02 \\
\textbf{40982} &
  0.71$\pm$0.02 &
  0.76$\pm$0.03 &
  0.79$\pm$0.02 &
  0.79$\pm$0.02 &
  0.79$\pm$0.02 &
  0.70$\pm$0.05 &
  0.75$\pm$0.02 &
  0.79$\pm$0.01 &
  0.72$\pm$0.02 &
  0.76$\pm$0.03 \\
\textbf{40994} &
  0.83$\pm$0.05 &
  0.49$\pm$0.03 &
  0.74$\pm$0.05 &
  0.73$\pm$0.05 &
  0.72$\pm$0.07 &
  0.51$\pm$0.07 &
  0.69$\pm$0.12 &
  0.82$\pm$0.05 &
  0.65$\pm$0.11 &
  0.68$\pm$0.10 \\ \bottomrule
\end{tabular}%
}
\end{table}

\begin{table}[H]
\vspace{-15pt}
\centering
\caption{For each considered model, we report the average Accuracy score and the corresponding standard deviation (across $10$ different seeds) on the $18$ benchmark datasets from the “OpenML-CC-18” benchmark suite.}
\label{tab:average_std_accuracy_score_small_tabular_data}
\scalebox{0.43}{%
\begin{tabular}{@{}c|cccccccccc@{}}
\toprule
\textbf{Dataset ID} &
  \multicolumn{10}{c}{\textbf{Model}} \\ \midrule
\textbf{} &
  \textbf{LogisticRegression} &
  \textbf{RandomForest} &
  \textbf{XGBoost} &
  \textbf{LightGBM} &
  \textbf{CatBoost} &
  \textbf{MLP} &
  \textbf{TabNet} &
  \textbf{TabPFN} &
  \textbf{\begin{tabular}[c]{@{}c@{}}HCNN\\ BootstrapNet\end{tabular}} &
  \textbf{\begin{tabular}[c]{@{}c@{}}HCNN\\ MeanSimMatrix\end{tabular}} \\ \midrule
\textbf{11} &
  0.88$\pm$0.02 &
  0.86$\pm$0.02 &
  0.97$\pm$0.01 &
  0.87$\pm$0.08 &
  0.90$\pm$0.02 &
  0.85$\pm$0.04 &
  0.89$\pm$0.02 &
  0.99$\pm$0.01 &
  0.98$\pm$0.01 &
  0.98$\pm$0.01 \\
\textbf{14} &
  0.79$\pm$0.01 &
  0.82$\pm$0.02 &
  0.81$\pm$0.02 &
  0.81$\pm$0.01 &
  0.82$\pm$0.02 &
  0.77$\pm$0.03 &
  0.82$\pm$0.01 &
  0.81$\pm$0.02 &
  0.81$\pm$0.01 &
  0.82$\pm$0.01 \\
\textbf{15} &
  0.97$\pm$0.01 &
  0.96$\pm$0.01 &
  0.96$\pm$0.01 &
  0.96$\pm$0.01 &
  0.97$\pm$0.01 &
  0.95$\pm$0.02 &
  0.96$\pm$0.02 &
  0.97$\pm$0.01 &
  0.97$\pm$0.01 &
  0.97$\pm$0.01 \\
\textbf{16} &
  0.95$\pm$0.01 &
  0.96$\pm$0.01 &
  0.95$\pm$0.01 &
  0.95$\pm$0.01 &
  0.96$\pm$0.01 &
  0.94$\pm$0.01 &
  0.96$\pm$0.01 &
  0.97$\pm$0.01 &
  0.94$\pm$0.01 &
  0.96$\pm$0.01 \\
\textbf{18} &
  0.74$\pm$0.02 &
  0.72$\pm$0.02 &
  0.71$\pm$0.02 &
  0.72$\pm$0.02 &
  0.72$\pm$0.02 &
  0.70$\pm$0.02 &
  0.74$\pm$0.01 &
  0.73$\pm$0.02 &
  0.74$\pm$0.02 &
  0.74$\pm$0.02 \\
\textbf{22} &
  0.82$\pm$0.02 &
  0.77$\pm$0.02 &
  0.79$\pm$0.01 &
  0.78$\pm$0.02 &
  0.79$\pm$0.02 &
  0.77$\pm$0.03 &
  0.82$\pm$0.02 &
  0.82$\pm$0.02 &
  0.82$\pm$0.02 &
  0.82$\pm$0.01 \\
\textbf{37} &
  0.76$\pm$0.03 &
  0.75$\pm$0.03 &
  0.74$\pm$0.04 &
  0.74$\pm$0.03 &
  0.73$\pm$0.03 &
  0.73$\pm$0.03 &
  0.73$\pm$0.05 &
  0.76$\pm$0.03 &
  0.75$\pm$0.04 &
  0.76$\pm$0.02 \\
\textbf{54} &
  0.76$\pm$0.03 &
  0.73$\pm$0.03 &
  0.76$\pm$0.04 &
  0.75$\pm$0.04 &
  0.76$\pm$0.04 &
  0.68$\pm$0.06 &
  0.80$\pm$0.04 &
  0.84$\pm$0.03 &
  0.82$\pm$0.04 &
  0.82$\pm$0.03 \\
\textbf{458} &
  1.00$\pm$0.00 &
  0.99$\pm$0.01 &
  0.99$\pm$0.00 &
  0.99$\pm$0.01 &
  0.99$\pm$0.01 &
  0.96$\pm$0.03 &
  0.97$\pm$0.02 &
  1.00$\pm$0.00 &
  0.98$\pm$0.02 &
  0.99$\pm$0.01 \\
\textbf{1049} &
  0.90$\pm$0.01 &
  0.90$\pm$0.02 &
  0.90$\pm$0.01 &
  0.90$\pm$0.02 &
  0.89$\pm$0.02 &
  0.88$\pm$0.01 &
  0.89$\pm$0.02 &
  0.91$\pm$0.01 &
  0.90$\pm$0.01 &
  0.90$\pm$0.01 \\
\textbf{1063} &
  0.82$\pm$0.03 &
  0.82$\pm$0.03 &
  0.80$\pm$0.03 &
  0.82$\pm$0.03 &
  0.81$\pm$0.02 &
  0.76$\pm$0.06 &
  0.80$\pm$0.02 &
  0.82$\pm$0.04 &
  0.82$\pm$0.04 &
  0.82$\pm$0.02 \\
\textbf{1068} &
  0.93$\pm$0.01 &
  0.94$\pm$0.01 &
  0.93$\pm$0.02 &
  0.93$\pm$0.01 &
  0.93$\pm$0.01 &
  0.88$\pm$0.08 &
  0.91$\pm$0.04 &
  0.93$\pm$0.01 &
  0.93$\pm$0.01 &
  0.93$\pm$0.01 \\
\textbf{1462} &
  0.98$\pm$0.01 &
  0.99$\pm$0.00 &
  0.99$\pm$0.00 &
  1.00$\pm$0.00 &
  1.00$\pm$0.00 &
  0.94$\pm$0.05 &
  0.97$\pm$0.04 &
  1.00$\pm$0.00 &
  1.00$\pm$0.00 &
  1.00$\pm$0.00 \\
\textbf{1464} &
  0.78$\pm$0.03 &
  0.79$\pm$0.02 &
  0.77$\pm$0.04 &
  0.78$\pm$0.02 &
  0.74$\pm$0.03 &
  0.76$\pm$0.05 &
  0.76$\pm$0.05 &
  0.79$\pm$0.03 &
  0.79$\pm$0.02 &
  0.79$\pm$0.02 \\
\textbf{1494} &
  0.87$\pm$0.01 &
  0.86$\pm$0.02 &
  0.85$\pm$0.02 &
  0.86$\pm$0.02 &
  0.85$\pm$0.02 &
  0.83$\pm$0.04 &
  0.83$\pm$0.07 &
  0.88$\pm$0.01 &
  0.85$\pm$0.02 &
  0.85$\pm$0.01 \\
\textbf{1510} &
  0.97$\pm$0.01 &
  0.95$\pm$0.02 &
  0.95$\pm$0.01 &
  0.95$\pm$0.02 &
  0.96$\pm$0.01 &
  0.93$\pm$0.02 &
  0.94$\pm$0.02 &
  0.97$\pm$0.01 &
  0.96$\pm$0.01 &
  0.96$\pm$0.02 \\
\textbf{40982} &
  0.71$\pm$0.02 &
  0.75$\pm$0.03 &
  0.77$\pm$0.02 &
  0.78$\pm$0.02 &
  0.78$\pm$0.01 &
  0.72$\pm$0.02 &
  0.73$\pm$0.01 &
  0.77$\pm$0.02 &
  0.71$\pm$0.02 &
  0.75$\pm$0.01 \\
\textbf{40994} &
  0.95$\pm$0.01 &
  0.91$\pm$0.02 &
  0.94$\pm$0.01 &
  0.93$\pm$0.01 &
  0.93$\pm$0.02 &
  0.84$\pm$0.11 &
  0.91$\pm$0.02 &
  0.95$\pm$0.01 &
  0.93$\pm$0.03 &
  0.93$\pm$0.02 \\ \bottomrule
\end{tabular}%
}
\end{table}

\begin{table}[H]
\vspace{-15pt}
\centering
\caption{For each considered model, we report the average MCC and the corresponding standard deviation (across $10$ different seeds) on the $18$ benchmark datasets from the “OpenML-CC-18” benchmark suite.}
\label{tab:average_std_mcc_score_small_tabular_data}
\scalebox{0.43}{%
\begin{tabular}{@{}c|cccccccccc@{}}
\toprule
\textbf{Dataset ID} &
  \multicolumn{10}{c}{\textbf{Model}} \\ \midrule
\textbf{} &
  \textbf{LogisticRegression} &
  \textbf{RandomForest} &
  \textbf{XGBoost} &
  \textbf{LightGBM} &
  \textbf{CatBoost} &
  \textbf{MLP} &
  \textbf{TabNet} &
  \textbf{TabPFN} &
  \textbf{\begin{tabular}[c]{@{}c@{}}HCNN\\ BootstrapNet\end{tabular}} &
  \textbf{\begin{tabular}[c]{@{}c@{}}HCNN\\ MeanSimMatrix\end{tabular}} \\ \midrule
\textbf{11} &
  0.79$\pm$0.03 &
  0.75$\pm$0.03 &
  0.95$\pm$0.02 &
  0.78$\pm$0.11 &
  0.83$\pm$0.04 &
  0.73$\pm$0.06 &
  0.80$\pm$0.04 &
  0.98$\pm$0.01 &
  0.96$\pm$0.02 &
  0.96$\pm$0.02 \\
\textbf{14} &
  0.77$\pm$0.02 &
  0.80$\pm$0.02 &
  0.79$\pm$0.02 &
  0.79$\pm$0.02 &
  0.80$\pm$0.02 &
  0.75$\pm$0.03 &
  0.80$\pm$0.01 &
  0.80$\pm$0.02 &
  0.79$\pm$0.02 &
  0.80$\pm$0.02 \\
\textbf{15} &
  0.93$\pm$0.02 &
  0.92$\pm$0.02 &
  0.92$\pm$0.02 &
  0.92$\pm$0.03 &
  0.92$\pm$0.03 &
  0.90$\pm$0.04 &
  0.91$\pm$0.04 &
  0.94$\pm$0.02 &
  0.93$\pm$0.02 &
  0.93$\pm$0.02 \\
\textbf{16} &
  0.94$\pm$0.01 &
  0.95$\pm$0.01 &
  0.95$\pm$0.01 &
  0.94$\pm$0.01 &
  0.96$\pm$0.01 &
  0.93$\pm$0.01 &
  0.95$\pm$0.01 &
  0.96$\pm$0.01 &
  0.94$\pm$0.01 &
  0.96$\pm$0.01 \\
\textbf{18} &
  0.71$\pm$0.02 &
  0.69$\pm$0.03 &
  0.68$\pm$0.03 &
  0.69$\pm$0.02 &
  0.68$\pm$0.03 &
  0.67$\pm$0.02 &
  0.71$\pm$0.01 &
  0.70$\pm$0.02 &
  0.72$\pm$0.03 &
  0.71$\pm$0.02 \\
\textbf{22} &
  0.80$\pm$0.02 &
  0.74$\pm$0.02 &
  0.76$\pm$0.01 &
  0.76$\pm$0.02 &
  0.77$\pm$0.02 &
  0.75$\pm$0.03 &
  0.80$\pm$0.03 &
  0.80$\pm$0.02 &
  0.80$\pm$0.02 &
  0.80$\pm$0.01 \\
\textbf{37} &
  0.47$\pm$0.08 &
  0.45$\pm$0.07 &
  0.44$\pm$0.09 &
  0.44$\pm$0.06 &
  0.42$\pm$0.06 &
  0.41$\pm$0.07 &
  0.40$\pm$0.10 &
  0.48$\pm$0.07 &
  0.45$\pm$0.09 &
  0.46$\pm$0.05 \\
\textbf{54} &
  0.68$\pm$0.05 &
  0.65$\pm$0.05 &
  0.68$\pm$0.06 &
  0.67$\pm$0.06 &
  0.69$\pm$0.06 &
  0.58$\pm$0.08 &
  0.74$\pm$0.05 &
  0.78$\pm$0.04 &
  0.76$\pm$0.05 &
  0.76$\pm$0.04 \\
\textbf{458} &
  0.99$\pm$0.01 &
  0.98$\pm$0.01 &
  0.99$\pm$0.01 &
  0.98$\pm$0.01 &
  0.98$\pm$0.01 &
  0.94$\pm$0.04 &
  0.96$\pm$0.02 &
  1.00$\pm$0.00 &
  0.96$\pm$0.03 &
  0.99$\pm$0.01 \\
\textbf{1049} &
  0.50$\pm$0.05 &
  0.45$\pm$0.07 &
  0.50$\pm$0.05 &
  0.49$\pm$0.10 &
  0.49$\pm$0.08 &
  0.32$\pm$0.16 &
  0.43$\pm$0.18 &
  0.57$\pm$0.07 &
  0.52$\pm$0.05 &
  0.51$\pm$0.08 \\
\textbf{1063} &
  0.39$\pm$0.11 &
  0.41$\pm$0.08 &
  0.32$\pm$0.17 &
  0.42$\pm$0.11 &
  0.42$\pm$0.06 &
  0.30$\pm$0.15 &
  0.25$\pm$0.15 &
  0.41$\pm$0.12 &
  0.40$\pm$0.12 &
  0.40$\pm$0.09 \\
\textbf{1068} &
  0.20$\pm$0.05 &
  0.28$\pm$0.12 &
  0.30$\pm$0.08 &
  0.30$\pm$0.10 &
  0.30$\pm$0.13 &
  0.07$\pm$0.10 &
  0.06$\pm$0.09 &
  0.09$\pm$0.13 &
  0.14$\pm$0.13 &
  0.25$\pm$0.08 \\
\textbf{1462} &
  0.96$\pm$0.01 &
  0.98$\pm$0.01 &
  0.99$\pm$0.01 &
  0.99$\pm$0.01 &
  1.00$\pm$0.01 &
  0.88$\pm$0.10 &
  0.94$\pm$0.08 &
  1.00$\pm$0.00 &
  1.00$\pm$0.00 &
  1.00$\pm$0.00 \\
\textbf{1464} &
  0.24$\pm$0.08 &
  0.33$\pm$0.09 &
  0.27$\pm$0.08 &
  0.30$\pm$0.06 &
  0.22$\pm$0.09 &
  0.19$\pm$0.13 &
  0.18$\pm$0.16 &
  0.33$\pm$0.07 &
  0.31$\pm$0.07 &
  0.31$\pm$0.08 \\
\textbf{1494} &
  0.70$\pm$0.03 &
  0.67$\pm$0.04 &
  0.67$\pm$0.04 &
  0.67$\pm$0.03 &
  0.65$\pm$0.04 &
  0.63$\pm$0.07 &
  0.60$\pm$0.20 &
  0.73$\pm$0.03 &
  0.66$\pm$0.04 &
  0.67$\pm$0.03 \\
\textbf{1510} &
  0.93$\pm$0.03 &
  0.88$\pm$0.04 &
  0.90$\pm$0.03 &
  0.89$\pm$0.04 &
  0.90$\pm$0.02 &
  0.86$\pm$0.04 &
  0.87$\pm$0.04 &
  0.93$\pm$0.02 &
  0.92$\pm$0.03 &
  0.92$\pm$0.03 \\
\textbf{40982} &
  0.62$\pm$0.03 &
  0.68$\pm$0.03 &
  0.70$\pm$0.03 &
  0.71$\pm$0.02 &
  0.72$\pm$0.02 &
  0.63$\pm$0.03 &
  0.65$\pm$0.02 &
  0.70$\pm$0.02 &
  0.62$\pm$0.03 &
  0.67$\pm$0.02 \\
\textbf{40994} &
  0.68$\pm$0.10 &
  0.03$\pm$0.10 &
  0.53$\pm$0.08 &
  0.51$\pm$0.09 &
  0.49$\pm$0.13 &
  0.07$\pm$0.11 &
  0.39$\pm$0.25 &
  0.65$\pm$0.11 &
  0.38$\pm$0.24 &
  0.43$\pm$0.18 \\ \bottomrule
\end{tabular}%
}
\end{table}

\newpage

\section{}\label{app:Appendix_Number_Of_Parameters}

\begin{figure}[h]
    \centering
    \subfigure[]{{\includegraphics[width=0.48\textwidth]{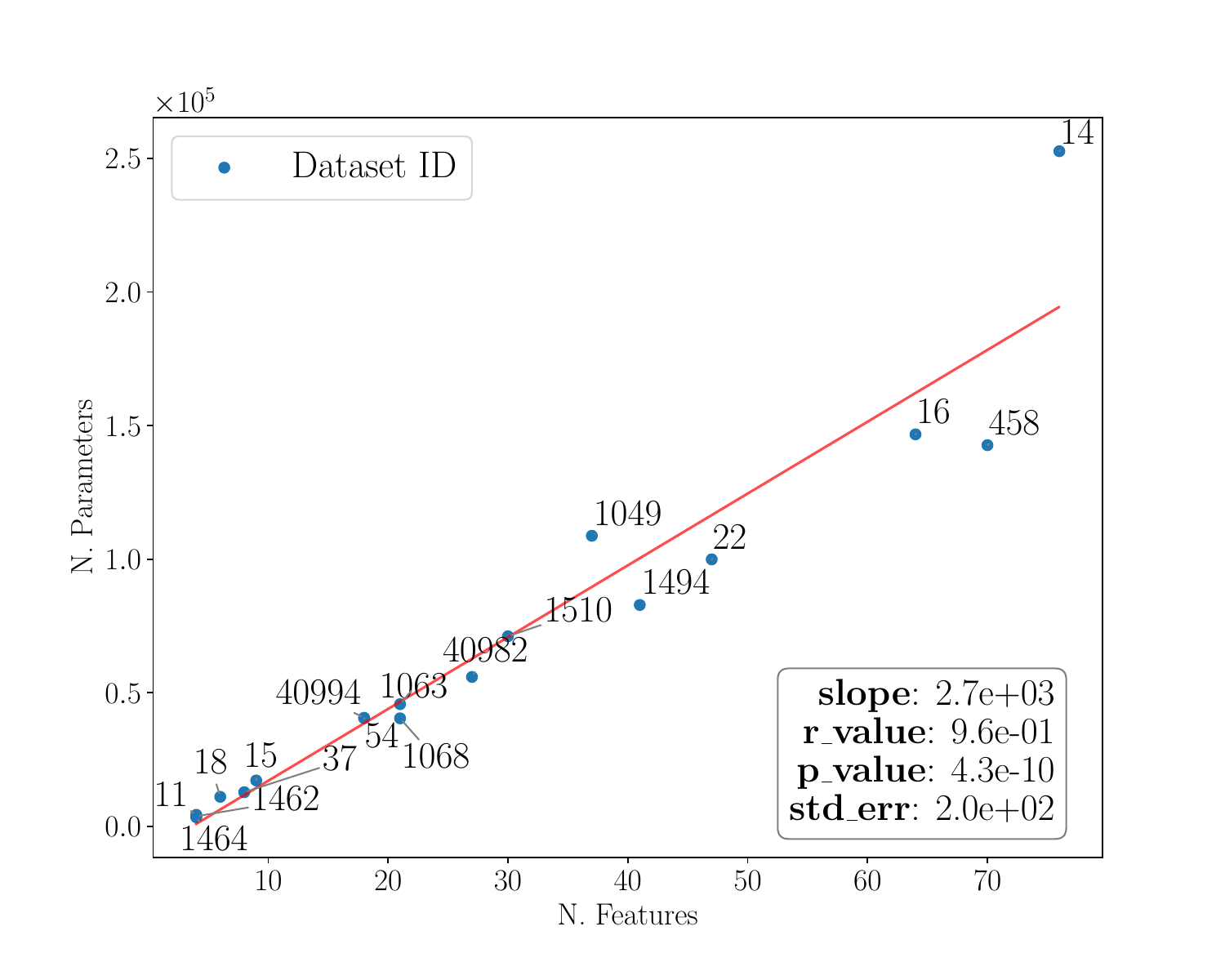}}\label{fig:HCNN_MeanSimMatrix_Parameters_Features}}
    \subfigure[]{{\includegraphics[width=0.48\textwidth]{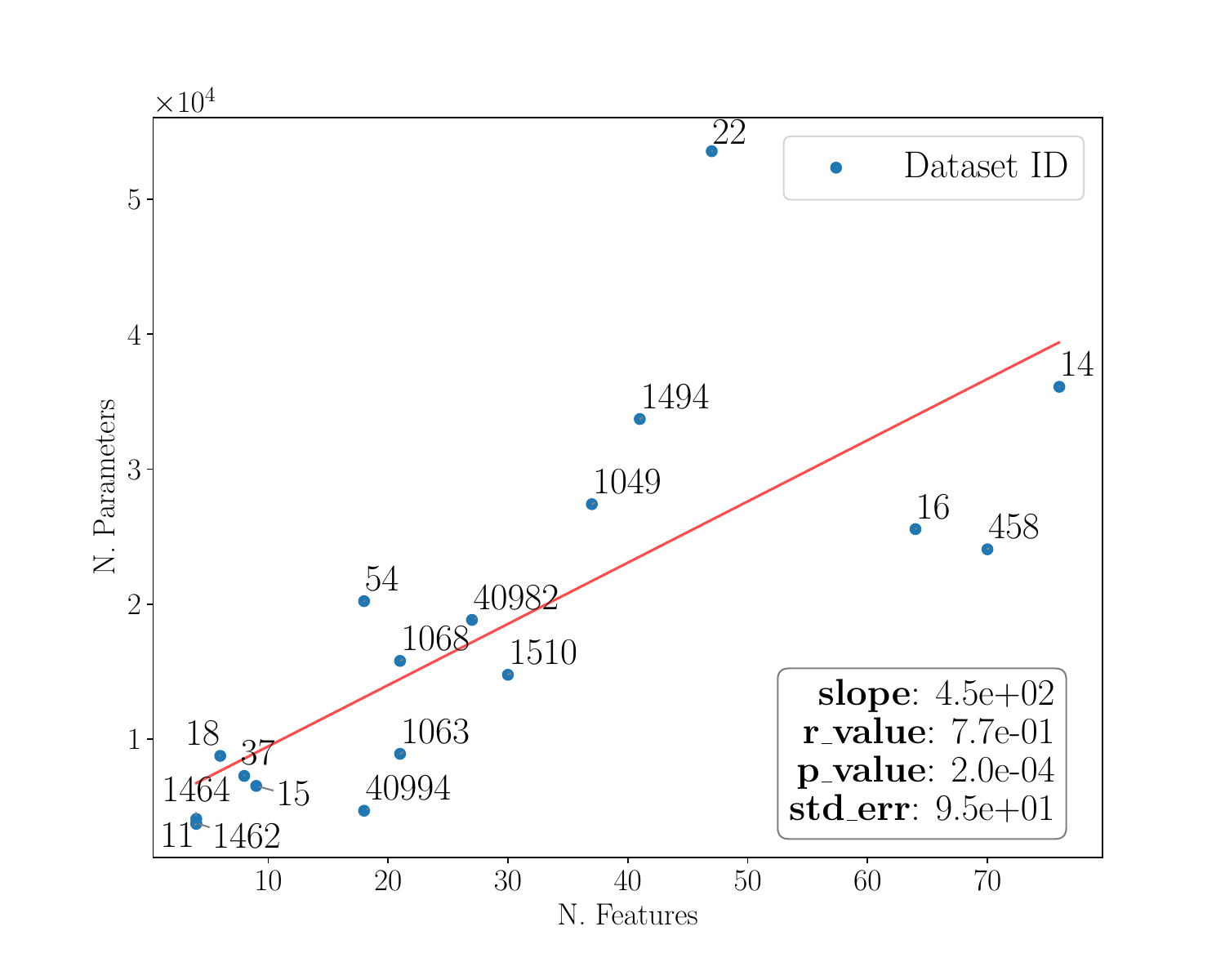}}\label{fig:HCNN_BootstrapNet_Parameters_Features}}
    \subfigure[]{{\includegraphics[width=0.5\textwidth]{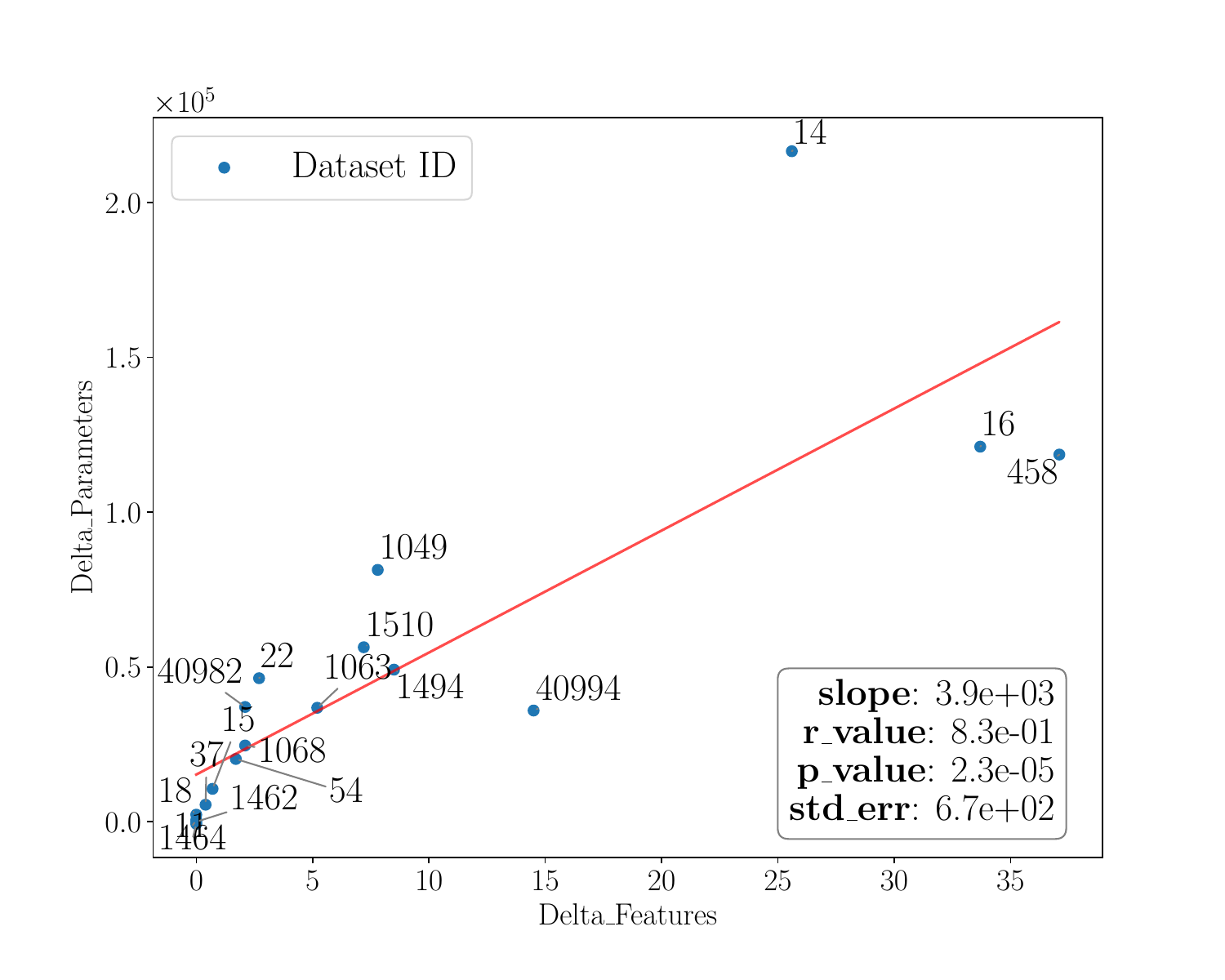}}\label{fig:HCNN_Delta_Comparison_Features}}
    \caption{Study of the relationship between the number of total features ($x$-axis) and the number of parameters ($y$-axis) for the (a) HCNN MeanSimMatrix and (b) HCNN BootstrapNet configuration. Figure (c) reports the relationship structure among the difference in the number of features ($x$-axis) and the difference in the number of total parameters ($y$-axis) when using the two above-mentioned configurations.}
    \label{fig:Parameters_Study}
\end{figure}

Figure \ref{fig:Parameters_Study} reports the relationship between the number of features and the total number of parameters for the \verb|HCNN| in its \verb|MeanSimMatrix| configuration, the relationship between the number of features and the total number of parameters for the \verb|HCNN| in its \verb|BootstrapNet| configuration and the relationship between the difference in the number of features and the difference in the total number of parameters in the two configurations. Looking at Figure \ref{fig:HCNN_MeanSimMatrix_Parameters_Features}, it is possible to conclude that a strong linear relationship exists between the number of features and the total number of parameters of the \verb|HCNN| model in the \verb|MeanSimMatrix| configuration. This finding was expected since the proposed model's architecture entirely depends on the complete homological structure of the underlying system. This means that, each time a new feature is introduced, we could potentially observe an increase in the number of edges, triangles, and tetrahedra, which in turn determines a proportional increase in the number of parameters of the \verb|HCNN| itself. On this point, we need to underline that the magnitude of the slope of the regression line heavily depends on the optimal hyper-parameters describing the number of filters in the two convolutional layers. Looking at Figure \ref{fig:HCNN_BootstrapNet_Parameters_Features}, we observe again a relatively strong linear relationship between the number of features and the total number of parameters of the \verb|HCNN| model in the \verb|BootstrapNet| configuration. The difference in r\_value between the two configurations is equal to $0.19$ and depends on the fact that, in the second case, the optimal threshold value which maximizes the model's performance (different across datasets), does not depend on the number of inputs features and determines the ablation of a number of features that has no dependence with any other factor. More generally, in the \verb|BootstrapNet| configuration, we observe a number of parameters that is, on average, one order of magnitude below the one in the \verb|HCNN MeanSimMatrix| configuration. To better study this finding, in Figure \ref{fig:HCNN_Delta_Comparison_Features} we report, on the $x$-axis the difference in the number of features $\Delta_f$ and on $y$-axis the difference in the number of parameters $\Delta_p$. As one can see, the linear relationship is strong only when the two deltas are low. For higher deltas, specifically for the three datasets “mfeat-fourier” (OpenML ID $14$), “mfeat-karhunen” (OpenML ID $16$), and “analcatdata\_authorship” (OpenML ID $458$), even if the decrement is significant for both parameters, the relationship is not linear.

\newpage

\section{}\label{app:Appendix_Results_Scalability_Datasets}

\begin{table}[H]
\vspace{-15pt}
\centering
\caption{For each considered model, we report the average F1\_score and the corresponding standard deviation (across $10$ different seeds) on the $9$ benchmark datasets from the “OpenML tabular
benchmark numerical classification” suite. The symbol "/" denotes the inability of the model to achieve convergence within the time or computing resources allocated for the corresponding learning task.}
\label{tab:average_std_f1_score_scalability_data}
\resizebox{\columnwidth}{!}{%
\begin{tabular}{@{}ccccccccccc@{}}
\toprule
\textbf{Dataset ID} &
  \multicolumn{10}{c}{\textbf{Model}} \\ \midrule
\multicolumn{1}{c|}{\textbf{}} &
  \textbf{LogisticRegression} &
  \textbf{RandomForest} &
  \textbf{XGBoost} &
  \textbf{LightGBM} &
  \textbf{CatBoost} &
  \textbf{MLP} &
  \textbf{TabNet} &
  \textbf{TabPFN} &
  \textbf{\begin{tabular}[c]{@{}c@{}}HCNN\\ BootstrapNet\end{tabular}} &
  \textbf{\begin{tabular}[c]{@{}c@{}}HCNN\\ MeanSimMatrix\end{tabular}} \\ \midrule
\multicolumn{1}{c|}{\textbf{361055}} &
  0.72$\pm$0.01 &
  0.78$\pm$0.00 &
  0.77$\pm$0.00 &
  0.78$\pm$0.00 &
  0.78$\pm$0.00 &
  0.75$\pm$0.01 &
  0.66$\pm$0.14 &
  / &
  0.77$\pm$0.01 &
  0.76$\pm$0.01 \\
\multicolumn{1}{c|}{\textbf{361062}} &
  0.86$\pm$0.01 &
  0.98$\pm$0.00 &
  0.98$\pm$0.00 &
  0.98$\pm$0.00 &
  0.98$\pm$0.00 &
  0.97$\pm$0.00 &
  0.98$\pm$0.00 &
  / &
  0.99$\pm$0.00 &
  0.99$\pm$0.00 \\
\multicolumn{1}{c|}{\textbf{361063}} &
  0.82$\pm$0.01 &
  0.88$\pm$0.01 &
  0.88$\pm$0.01 &
  0.88$\pm$0.01 &
  0.88$\pm$0.00 &
  0.86$\pm$0.00 &
  0.86$\pm$0.01 &
  / &
  0.88$\pm$0.01 &
  0.88$\pm$0.01 \\
\multicolumn{1}{c|}{\textbf{361065}} &
  0.77$\pm$0.01 &
  0.86$\pm$0.00 &
  0.86$\pm$0.01 &
  0.86$\pm$0.01 &
  0.86$\pm$0.00 &
  0.84$\pm$0.01 &
  0.86$\pm$0.01 &
  / &
  0.82$\pm$0.03 &
  0.85$\pm$0.00 \\
\multicolumn{1}{c|}{\textbf{361066}} &
  0.74$\pm$0.01 &
  0.80$\pm$0.01 &
  0.80$\pm$0.01 &
  0.80$\pm$0.01 &
  0.80$\pm$0.01 &
  0.78$\pm$0.01 &
  0.79$\pm$0.01 &
  / &
  0.80$\pm$0.01 &
  0.79$\pm$0.00 \\
\multicolumn{1}{c|}{\textbf{361275}} &
  0.67$\pm$0.01 &
  0.71$\pm$0.01 &
  0.71$\pm$0.01 &
  0.71$\pm$0.01 &
  0.71$\pm$0.01 &
  0.70$\pm$0.01 &
  0.70$\pm$0.01 &
  / &
  0.70$\pm$0.01 &
  0.70$\pm$0.01 \\
\multicolumn{1}{c|}{\textbf{361276}} &
  0.73$\pm$0.02 &
  0.78$\pm$0.01 &
  0.78$\pm$0.02 &
  0.78$\pm$0.02 &
  0.78$\pm$0.02 &
  0.75$\pm$0.02 &
  0.71$\pm$0.03 &
  / &
  0.72$\pm$0.02 &
  / \\
\multicolumn{1}{c|}{\textbf{361277}} &
  0.83$\pm$0.01 &
  0.89$\pm$0.00 &
  0.90$\pm$0.00 &
  0.90$\pm$0.00 &
  0.90$\pm$0.00 &
  0.86$\pm$0.01 &
  0.84$\pm$0.02 &
  / &
  0.88$\pm$0.01 &
  0.88$\pm$0.00 \\
\multicolumn{1}{c|}{\textbf{361278}} &
  0.71$\pm$0.01 &
  0.72$\pm$0.01 &
  0.72$\pm$0.01 &
  0.72$\pm$0.01 &
  0.72$\pm$0.01 &
  0.71$\pm$0.01 &
  0.71$\pm$0.01 &
  / &
  0.72$\pm$0.01 &
  0.72$\pm$0.01 \\ \bottomrule
\end{tabular}%
}
\end{table}

\begin{table}[H]
\vspace{-15pt}
\centering
\caption{For each considered model, we report the average Accuracy score and the corresponding standard deviation (across $10$ different seeds) on the $9$ benchmark datasets from the “OpenML tabular
benchmark numerical classification” suite. The symbol "/" denotes the inability of the model to achieve convergence within the time or computing resources allocated for the corresponding learning task.}
\label{tab:average_std_accuracy_score_scalability_data}
\resizebox{\columnwidth}{!}{%
\begin{tabular}{@{}ccccccccccc@{}}
\toprule
\textbf{Dataset ID} &
  \multicolumn{10}{c}{\textbf{Model}} \\ \midrule
\multicolumn{1}{c|}{\textbf{}} &
  \textbf{LogisticRegression} &
  \textbf{RandomForest} &
  \textbf{XGBoost} &
  \textbf{LightGBM} &
  \textbf{CatBoost} &
  \textbf{MLP} &
  \textbf{TabNet} &
  \textbf{TabPFN} &
  \textbf{\begin{tabular}[c]{@{}c@{}}HCNN\\ BootstrapNet\end{tabular}} &
  \textbf{\begin{tabular}[c]{@{}c@{}}HCNN\\ MeanSimMatrix\end{tabular}} \\ \midrule
\multicolumn{1}{c|}{\textbf{361055}} &
  0.72$\pm$0.01 &
  0.78$\pm$0.00 &
  0.77$\pm$0.00 &
  0.78$\pm$0.00 &
  0.78$\pm$0.00 &
  0.75$\pm$0.01 &
  0.69$\pm$0.09 &
  / &
  0.77$\pm$0.01 &
  0.76$\pm$0.01 \\
\multicolumn{1}{c|}{\textbf{361062}} &
  0.86$\pm$0.01 &
  0.98$\pm$0.00 &
  0.98$\pm$0.00 &
  0.98$\pm$0.00 &
  0.98$\pm$0.00 &
  0.97$\pm$0.00 &
  0.98$\pm$0.00 &
  / &
  0.99$\pm$0.00 &
  0.99$\pm$0.00 \\
\multicolumn{1}{c|}{\textbf{361063}} &
  0.82$\pm$0.01 &
  0.88$\pm$0.01 &
  0.88$\pm$0.01 &
  0.88$\pm$0.01 &
  0.88$\pm$0.00 &
  0.86$\pm$0.00 &
  0.86$\pm$0.01 &
  / &
  0.88$\pm$0.01 &
  0.88$\pm$0.01 \\
\multicolumn{1}{c|}{\textbf{361065}} &
  0.77$\pm$0.01 &
  0.86$\pm$0.00 &
  0.86$\pm$0.01 &
  0.86$\pm$0.01 &
  0.86$\pm$0.00 &
  0.84$\pm$0.01 &
  0.86$\pm$0.01 &
  / &
  0.82$\pm$0.03 &
  0.85$\pm$0.00 \\
\multicolumn{1}{c|}{\textbf{361066}} &
  0.75$\pm$0.01 &
  0.80$\pm$0.01 &
  0.80$\pm$0.01 &
  0.80$\pm$0.01 &
  0.80$\pm$0.01 &
  0.78$\pm$0.01 &
  0.79$\pm$0.01 &
  / &
  0.80$\pm$0.01 &
  0.79$\pm$0.00 \\
\multicolumn{1}{c|}{\textbf{361275}} &
  0.67$\pm$0.01 &
  0.71$\pm$0.01 &
  0.71$\pm$0.01 &
  0.71$\pm$0.01 &
  0.71$\pm$0.01 &
  0.70$\pm$0.01 &
  0.70$\pm$0.01 &
  / &
  0.71$\pm$0.01 &
  0.71$\pm$0.01 \\
\multicolumn{1}{c|}{\textbf{361276}} &
  0.73$\pm$0.02 &
  0.78$\pm$0.01 &
  0.78$\pm$0.02 &
  0.78$\pm$0.02 &
  0.78$\pm$0.02 &
  0.75$\pm$0.02 &
  0.71$\pm$0.03 &
  / &
  0.73$\pm$0.02 &
  / \\
\multicolumn{1}{c|}{\textbf{361277}} &
  0.83$\pm$0.01 &
  0.89$\pm$0.00 &
  0.90$\pm$0.00 &
  0.90$\pm$0.00 &
  0.90$\pm$0.00 &
  0.86$\pm$0.01 &
  0.84$\pm$0.02 &
  / &
  0.88$\pm$0.01 &
  0.88$\pm$0.00 \\
\multicolumn{1}{c|}{\textbf{361278}} &
  0.71$\pm$0.01 &
  0.72$\pm$0.01 &
  0.72$\pm$0.01 &
  0.72$\pm$0.01 &
  0.72$\pm$0.01 &
  0.71$\pm$0.01 &
  0.71$\pm$0.01 &
  / &
  0.72$\pm$0.01 &
  0.72$\pm$0.01 \\ \bottomrule
\end{tabular}%
}
\end{table}

\begin{table}[H]
\vspace{-15pt}
\centering
\caption{For each considered model, we report the average MCC and the corresponding standard deviation (across $10$ different seeds) on the $9$ benchmark datasets from the “OpenML tabular
benchmark numerical classification” suite. The symbol "/" denotes the inability of the model to achieve convergence within the time or computing resources allocated for the corresponding learning task.}
\label{tab:average_std_mcc_score_scalability_data}
\resizebox{\columnwidth}{!}{%
\begin{tabular}{@{}ccccccccccc@{}}
\toprule
\textbf{Dataset ID} &
  \multicolumn{10}{c}{\textbf{Model}} \\ \midrule
\multicolumn{1}{c|}{\textbf{}} &
  \textbf{LogisticRegression} &
  \textbf{RandomForest} &
  \textbf{XGBoost} &
  \textbf{LightGBM} &
  \textbf{CatBoost} &
  \textbf{MLP} &
  \textbf{TabNet} &
  \textbf{TabPFN} &
  \textbf{\begin{tabular}[c]{@{}c@{}}HCNN\\ BootstrapNet\end{tabular}} &
  \textbf{\begin{tabular}[c]{@{}c@{}}HCNN\\ MeanSimMatrix\end{tabular}} \\ \midrule
\multicolumn{1}{c|}{\textbf{361055}} &
  0.45$\pm$0.03 &
  0.56$\pm$0.01 &
  0.55$\pm$0.01 &
  0.55$\pm$0.01 &
  0.55$\pm$0.00 &
  0.50$\pm$0.02 &
  0.41$\pm$0.15 &
  / &
  0.53$\pm$0.02 &
  0.52$\pm$0.02 \\
\multicolumn{1}{c|}{\textbf{361062}} &
  0.72$\pm$0.01 &
  0.96$\pm$0.01 &
  0.96$\pm$0.00 &
  0.97$\pm$0.00 &
  0.97$\pm$0.00 &
  0.95$\pm$0.00 &
  0.96$\pm$0.01 &
  / &
  0.97$\pm$0.00 &
  0.97$\pm$0.00 \\
\multicolumn{1}{c|}{\textbf{361063}} &
  0.65$\pm$0.02 &
  0.76$\pm$0.01 &
  0.76$\pm$0.01 &
  0.77$\pm$0.01 &
  0.76$\pm$0.01 &
  0.73$\pm$0.01 &
  0.72$\pm$0.02 &
  / &
  0.75$\pm$0.01 &
  0.75$\pm$0.01 \\
\multicolumn{1}{c|}{\textbf{361065}} &
  0.55$\pm$0.01 &
  0.72$\pm$0.01 &
  0.72$\pm$0.01 &
  0.72$\pm$0.01 &
  0.72$\pm$0.01 &
  0.69$\pm$0.01 &
  0.72$\pm$0.01 &
  / &
  0.64$\pm$0.05 &
  0.71$\pm$0.01 \\
\multicolumn{1}{c|}{\textbf{361066}} &
  0.49$\pm$0.01 &
  0.60$\pm$0.01 &
  0.60$\pm$0.02 &
  0.60$\pm$0.01 &
  0.60$\pm$0.02 &
  0.56$\pm$0.02 &
  0.57$\pm$0.01 &
  / &
  0.59$\pm$0.01 &
  0.59$\pm$0.01 \\
\multicolumn{1}{c|}{\textbf{361275}} &
  0.34$\pm$0.02 &
  0.43$\pm$0.02 &
  0.42$\pm$0.01 &
  0.42$\pm$0.01 &
  0.42$\pm$0.02 &
  0.41$\pm$0.01 &
  0.42$\pm$0.02 &
  / &
  0.42$\pm$0.01 &
  0.42$\pm$0.01 \\
\multicolumn{1}{c|}{\textbf{361276}} &
  0.47$\pm$0.03 &
  0.56$\pm$0.02 &
  0.57$\pm$0.03 &
  0.55$\pm$0.04 &
  0.55$\pm$0.03 &
  0.50$\pm$0.05 &
  0.43$\pm$0.06 &
  / &
  0.45$\pm$0.04 &
  / \\
\multicolumn{1}{c|}{\textbf{361277}} &
  0.66$\pm$0.02 &
  0.78$\pm$0.01 &
  0.81$\pm$0.01 &
  0.81$\pm$0.01 &
  0.81$\pm$0.01 &
  0.72$\pm$0.02 &
  0.69$\pm$0.04 &
  / &
  0.76$\pm$0.02 &
  0.76$\pm$0.01 \\
\multicolumn{1}{c|}{\textbf{361278}} &
  0.42$\pm$0.02 &
  0.44$\pm$0.01 &
  0.44$\pm$0.02 &
  0.44$\pm$0.02 &
  0.44$\pm$0.02 &
  0.43$\pm$0.02 &
  0.42$\pm$0.02 &
  / &
  0.43$\pm$0.02 &
  0.44$\pm$0.02 \\ \bottomrule
\end{tabular}%
}
\end{table}

\end{document}